\documentclass[10pt,twocolumn,letterpaper]{article}

\usepackage{cvpr}
\usepackage{times}
\usepackage{epsfig}
\usepackage{graphicx}
\usepackage{amsmath}
\usepackage{amssymb}
\usepackage{color, colortbl}
\usepackage{multirow}
\usepackage{pifont}

\usepackage[dvipsnames]{xcolor}
\usepackage{comment}

\usepackage{textcomp}

\usepackage{makecell}

\usepackage[pagebackref=true,breaklinks=true,letterpaper=true,colorlinks,bookmarks=false]{hyperref}

\cvprfinalcopy 

\newcommand\rel[1]{\texttt{#1}}
\newcommand{\labelCount}{534 }
\newcommand{\nameDataset}{3DSSG}
\newcommand{\instanceCount}{48k } 
\newcommand{\numberEdges}{544k } 
\newcommand{\numberFrames}{363k } 
\newcommand{\relCount}{40} 
\newcommand{\numAttributesTotal}{48k } 
\newcommand{\numberAttributes}{93} 
\newcommand{\numberAttributesInstances}{21k } 

\newcommand{\npnet}{ObjPointNet }
\newcommand{\epnet}{RelPointNet }

\newcommand\blfootnote[1]{%
	\begingroup
	\renewcommand\thefootnote{}\footnote{#1}%
	\addtocounter{footnote}{-1}%
	\endgroup
}

\begin{document}

\title{Learning 3D Semantic Scene Graphs from 3D Indoor Reconstructions}

\author{Johanna Wald $^{1,}$ \thanks{the authors contributed equally to this paper}\\
\and
Helisa Dhamo $^{1,}$ \footnotemark[1]\\
\and
Nassir Navab $^{1}$\\
\and
Federico Tombari $^{1,2}$\\
\and
$^{1}$ Technische Universit\"at M\"unchen
\and
$^{2}$ Google 
}

\maketitle

\begin{abstract}
Scene understanding has been of high interest in computer vision. It encompasses not only identifying objects in a scene, but also their relationships within the given context. With this goal, a recent line of works tackles 3D semantic segmentation and scene layout prediction. In our work we focus on scene graphs, a data structure that organizes the entities of a scene in a graph, where objects are nodes and their relationships modeled as edges. We leverage inference on scene graphs as a way to carry out 3D scene understanding, mapping objects and their relationships. In particular, we propose a learned method that regresses a scene graph from the point cloud of a scene. Our novel architecture is based on PointNet and Graph Convolutional Networks (GCN). In addition, 
we introduce \nameDataset, a semi-automatically generated dataset, that contains semantically rich scene graphs of 3D scenes.
We show the application of our method in a domain-agnostic retrieval task, where graphs serve as an intermediate representation for 3D-3D and 2D-3D matching.
\end{abstract}

\section{Introduction}

3D scene understanding relates to the perception and interpretation of a scene from 3D data, with a focus on its semantic and geometric nature, which includes not only recognizing and localizing the objects present in the 3D space therein, but also their context and relationships. This thorough understanding is of high interest for various applications such as robotic navigation, augmented and virtual reality. Current 3D scene understanding works include perception tasks such as instance segmentation \cite{Hou2018, Lahoud_2019_ICCV, Thomas_2019_ICCV, Yi_2019_CVPR}, semantic segmentation \cite{Ruizhongtai2016, Ruizhongtai2017, Dai2018, Rethage2018} as well as 3D object detection and classification \cite{Song2015, Ruizhongtai2016, Qi2016, Zhou2017}. While these works mostly focus on object semantics, their context and relationships are primarily used to improve the per-object class accuracy.

Scene understanding from images has recently explored the use of scene graphs to aid understanding object relationships in addition to characterizing objects individually. Before that, scene graphs have been used in computer graphics to arrange spatial representations of a graphical scene, where nodes commonly represent scene entities (object instances), while the edges represent relative transformations between two nodes. This is a flexible representation of a scene which encompasses also complex spatial relations and operation grouping.  Some of these concepts where successively adapted or extended in computer vision datasets, such as support structures~\cite{Silberman2012}, semantic relationships and attributes~\cite{krishna2017visual} and hierarchical mapping of scene entities~\cite{armeni_iccv19}. Scene graphs have been shown to be relevant, for instance, for partial \cite{Wang14} and full matching \cite{johnson15} in image search, as well as image generation \cite{johnson2018image}. 

\begin{figure}[t]
\begin{center}
   \includegraphics[width=\linewidth]{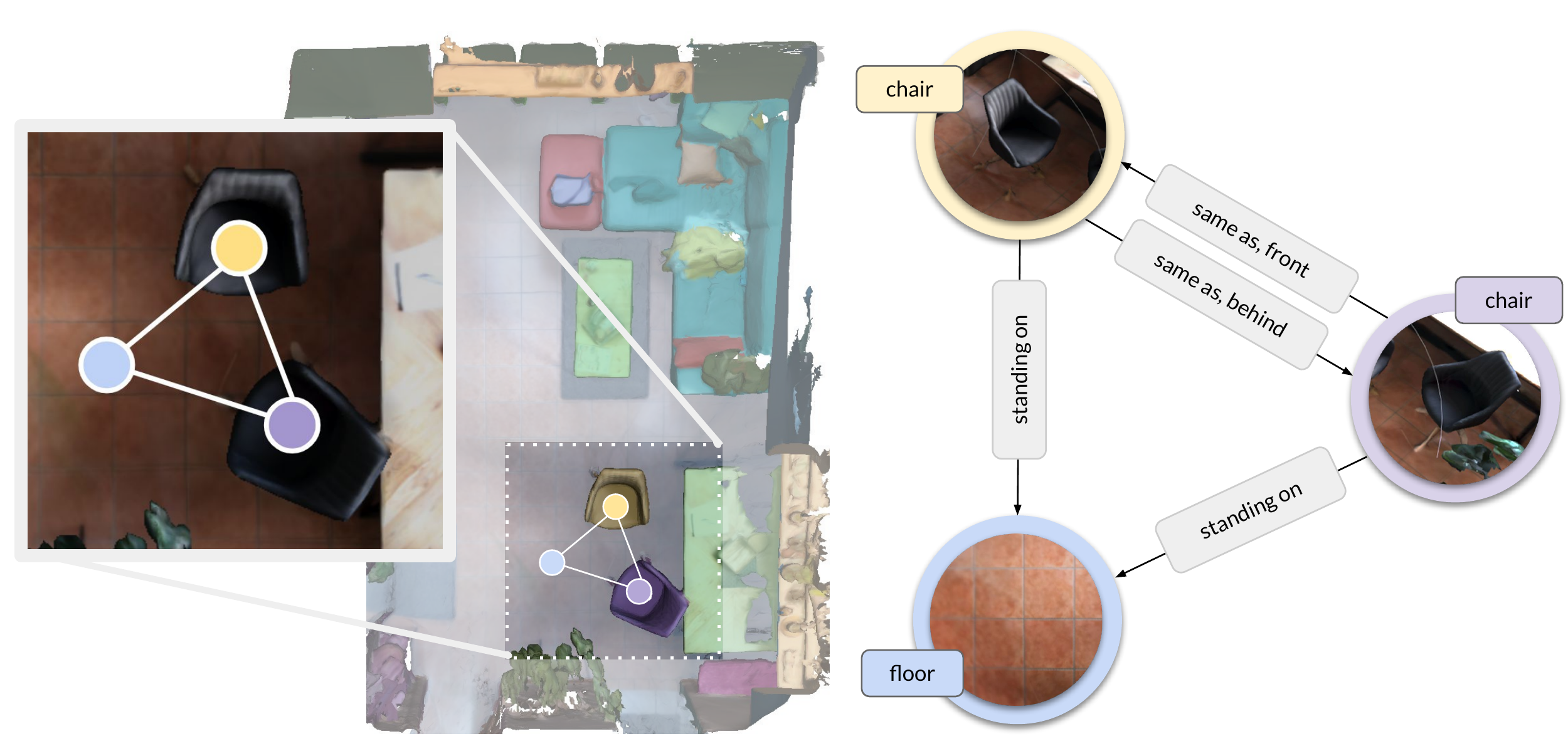}
\end{center}
   \caption{\textbf{Overview} 
   Given a class-agnostic instance segmentation of a 3D scene (left) our graph prediction network infers a semantic scene graph $\mathcal{G}$ (right) from a point cloud.}
\label{fig:teaser}
\end{figure}

In 3D, scene graphs have only recently gained more popularity~\cite{armeni_iccv19}. In this work, we want to focus on the \emph{semantic} aspects of 3D scene graphs as well as their potential. Our goal is to obtain dense graphs with labeled instances (nodes), semantically meaningful relationships (edges) such as \rel{lying on} or \rel{same as} and attributes including \textit{color, shape or affordances} (see Fig. \ref{fig:teaser}). These resemble the scene graph representation of~\cite{johnson15}, associated with images. We believe semantic scene graphs are especially important in 3D since
a) they are a compact representation, that describes a (potentially large) 3D scene,
b) they are robust towards small scene changes and noise and c) they close the gap between different domains, such as text or images. These properties make them suitable for cross domain tasks such as 2D-3D Scene Retrieval or VQA. 

We believe that the capability of regressing the scene graph of a given 3D scene can be a fundamental piece for 3D scene understanding, as a way to learn and represent object relationships and contextual information of an environment. For this purpose, we propose a learned method, based on PointNet \cite{Ruizhongtai2016} and Graph Convolutional Networks (GCNs) \cite{kipf2017semi}, to predict 3D semantic graphs. Given a class-agnostic instance segmentation of a 3D point cloud, we jointly infer a 3D scene graph composed of nodes (scene components) and edges (their relationships). For this purpose, we introduce a 3D semantic scene graph dataset that features detailed semantics in the nodes (instances) including attributes and edges (relationships), which will be publicly released\footnote{https://3DSSG.github.io}. Generating 3D semantic scene graphs from real-world scans is particularly challenging due to missing data and clutter and the complexity of the relationships between objects. For instance, two chairs that are of the \rel{same style} could have very different appearances, while a jacket \rel{lying on} one of them might occlude most of its visible surface. While our method outperforms the baseline, it operates end-to-end and is able to predict multiple relationships per edges. We further show how -- in a cross-domain scenario -- scene graphs serve as a common encoding between 3D and 2D in a scene retrieval task in changing conditions. Given a single image the task is to find the matching 3D model from a pool of scans. Scene graphs suit particularly well because they are inherently robust towards dynamic environments, which manifest illumination changes and \mbox{(non-)rigid} changes introduced by human activity. In summary, we explore the prediction and application of semantic scene graphs in 3D indoor environments with the following contributions:
\begin{itemize}
    \item We present \nameDataset, a large scale 3D dataset that extends 3RScan~\cite{Wald2019RIO} with semantic scene graph annotations, containing relationships, attributes and class hierarchies. Interestingly, 2D scene graphs can be obtained by rendering the 3D graphs, which results in \numberFrames graph-image pairs.
    \item We propose the first learned method that generates a semantic scene graph from a 3D point cloud.
    \item We show how 3D semantic scene graphs can be used in cross-domain retrieval, specifically 2D-3D scene retrieval of changing indoor environments. 
\end{itemize}

\section{Related Work}

\begin{figure*}[ht]
    \centering
    \includegraphics[width=\linewidth]{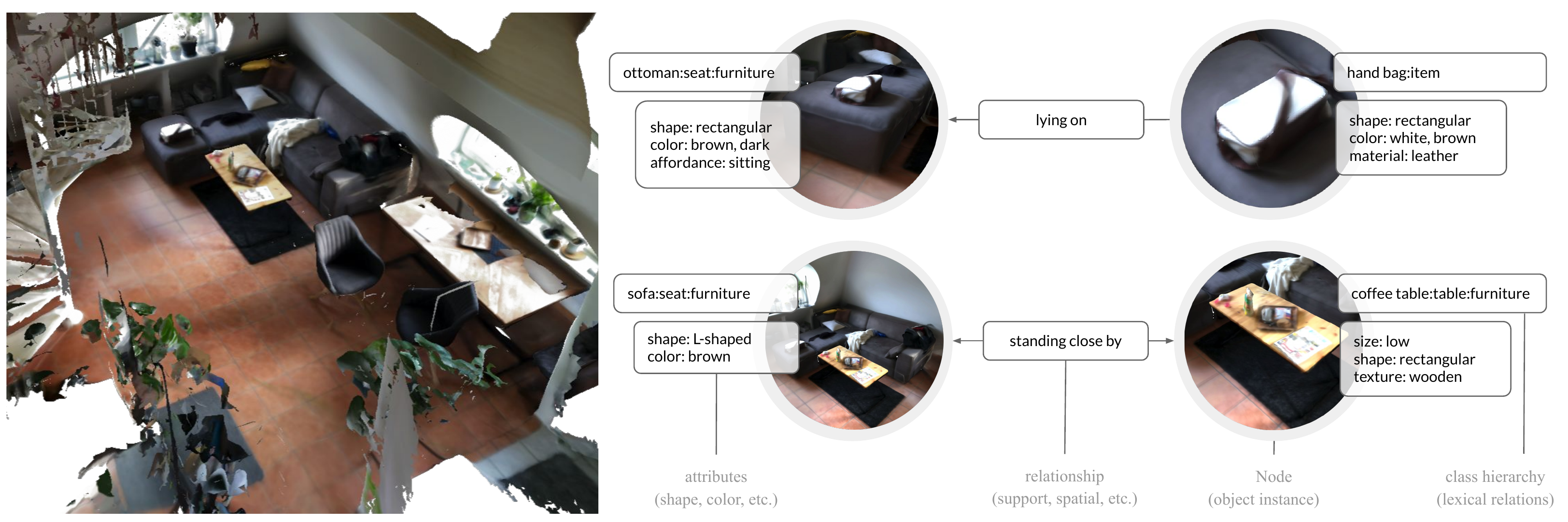}
    \caption{\textbf{Scene graph representation in \nameDataset{ }}including hierarchical class labels $c$ and attributes $A$ per node, as well as relationship triplets between nodes.}
    \label{fig:triples}
\end{figure*}

\paragraph{Semantic Scene Graphs with Images.} 

Johnson \etal \cite{johnson15} introduced scene graphs -- motivated by image retrieval -- as a representation that semantically describes an image, where each node is an object while edges represent interactions between them. Additionally, the object nodes contain attributes that describe object properties.
Later, Visual Genome \cite{krishna2017visual}, a large scale dataset with scene graph annotations on images, gave rise to a line of deep learning based advances on scene graph prediction from images~\cite{xu2017scenegraph,herzig2018mapping,qi2018attentive,zellers2018neural,li2017scene,yang2018graph,li2018factorizable,newell2017pixels}. These methods propose diverse strategies for graph estimation and processing, such as message passing~\cite{xu2017scenegraph}, graph convolutional networks (GCN)~\cite{yang2018graph}, permutation invariant architectures~\cite{herzig2018mapping} and attention mechanisms~\cite{qi2018attentive}. Most of these methods rely on an object detector to extract node- and edge-specific features prior to the graph computation \cite{xu2017scenegraph, yang2018graph,li2018factorizable}.
Recent works explore the reverse problem of using scene graphs to generate new images \cite{johnson2018image} or manipulate existing images \cite{Dhamo2020cvpr}. 

\paragraph{3D Understanding: From Objects to Relationships.}

An active research area within 3D scene understanding focuses on 3D semantic segmentation \cite{Ruizhongtai2016, Dai2017, Ruizhongtai2017, Engelmann_2017_ICCV, Rethage2018, Hou2018} and object detection and classification \cite{su15mvcnn, Ruizhongtai2016, Qi2016, Zhao_2019_CVPR}. These works mostly focus on object semantics and context is only used to improve object class accuracy. Holistic scene understanding \cite{Song2015} on the other side predicts not only object semantics, but also the scene layout and sometimes even the camera pose \cite{Siyuan_NIPS2018}. Scene context is often represented through a hierarchical tree, where the leaves are typically objects and the intermediate nodes group the objects in scene components or functional entities. A line of works use probabilistic grammar to parse scenes \cite{Liu:2014:CCS,NIPS2011_4236} or control scene synthesis \cite{Jiangijcv18}. Shi \etal~\cite{shi2019hierarchy} show that the object detection task benefits from joint prediction of the hierarchical context. GRAINS~\cite{Li2018grains} explore hierarchical graphs to synthesize diverse 3D scenes, using a recursive VAE that generates a layout, followed by object retrieval. 
In a 3D from single image scenario, Kulkarni \etal~\cite{kulkarni20193d} consider relative 3D poses between objects (as edges), which are shown to outperform neighbour-agnostic 6D pose estimation.

Another line of works incorporate graphs structures for object-level understanding, rather than entire scenes. Te \etal~\cite{Te2018} use a Graph CNN for semantic segmentation of object parts. StructureNet~\cite{Mo2019StructureNet} represent the latent space of an object as a hierarchical graph of composing parts, with the goal of generating plausible shapes. However, all of these works are either focused on object parts, or do not consider semantic relationships that go beyond generic edges (without semantic labels) or relative transformations. In the context of semantic scene graphs on synthetic data, Fisher \etal~\cite{Fisher11characterizingstructural} use graph kernels for 3D scene comparison, based on support and spatial relationships. Ma \etal~\cite{Ma2018language} parse natural language into semantic scene graphs, considering pairwise and group relationships, to progressively retrieve sub-scenes for 3D synthesis. 

Only recently the community started to explore semantic relationships in 3D and on real world data.
Armeni \etal~\cite{armeni_iccv19} present a hierarchical mapping of 3D models of large spaces in four layers: camera, object, room and building. While they feature smaller graphs (see Tbl. \ref{table:dataset_comparision}) their focus is not on semantically meaningful inter-instance relationships such as support. Moreover, the absence of changing scenes, does not enable the proposed 3D scene retrieval task.

\paragraph{3D Scene Retrieval} 

Many image-based 3D retrieval works focus on retrieving 3D CAD models from RGB images: IM2CAD generates a 3D scene from a single image by detecting the objects, estimating the room layout and retrieving a corresponding CAD model for each bounding box \cite{izadinia2017im2cad}. Pix3D on the other hand propose a dataset for single image 3D shape modeling based on highly accurate 3D model alignments in the 2D images \cite{Sun2018}. Liu~\etal show improved 2D-3D model retrieval by simulating local context to generate false occlusion \cite{Liu2018}. The SHREC benchmark \cite{Rashid2018, Rashid2019}, enables 2D-3D retrieval of diverse scenes (beach, bedroom or castle), while \cite{Ma2018language} and \cite{Fisher11characterizingstructural} operate on indoor environments but also only focus on synthetic data rather than real 3D reconstructions. 
\section{3D Semantic Scene Graphs}
\label{sec:data}
\blfootnote{* We compare against the 3D scene graph dataset on the tiny Gibson split, the most recent release at the time of the submission}

With this work, we release \nameDataset{ }which provides 3D semantic scene graphs for 3RScan~\cite{Wald2019RIO}, a large scale, real-world dataset which features 1482 3D reconstructions of 478 naturally changing indoor environments. 
A semantic scene graph $\mathcal{G}$ in \nameDataset, is a set of tuples (\( \mathcal{N, R} \)) between nodes \( \mathcal{N} \) and edges \( \mathcal{R} \) (see Fig. \ref{fig:triples}). Nodes represent specific 3D object instances in a 3D scan. In contrast to previous works \cite{krishna2017visual, armeni_iccv19, Dai2017, Wald2019RIO}, our nodes are not assigned a single object category \( \mathcal{C} \) only, but instead are defined by a hierarchy of classes  $c = (c_1, ..., c_d)$ where $c \in \mathcal{C}^d$, and $d$ can vary. Additionally to these object categories each node has a set of attributes $A$ that describe the visual and physical appearance of the object instance. A special subset of the attributes are affordances \cite{xiazamirhe2018gibsonenv}. We consider them particularly important since we deal with changing environments. The edges in our graphs define semantic relationships (predicates) between the nodes such as \rel{standing on, hanging on, more comfortable than, same material}. 
To obtain the data in \nameDataset{ }we combine semantic annotations with geometric data and additional human verification to ensure high quality graphs. In summary, our dataset features 1482 scene graphs with \instanceCount object nodes and \numberEdges edges. An interesting feature of 3D scene graphs is that they can easily be \textit{rendered} to 2D. Given a 3D model and a camera pose, one can filter the graph nodes and edges that are present in that image. Support and attribute comparison relations remain the same, while directional relationships (\rel{left, right, behind, front}) must be updated automatically for the new viewpoint. Given the \numberFrames RGB-D images with camera poses of 3RScan, this results in \numberFrames 2D scene graphs. A comparison of our dataset with the only other real 3D semantic scene graph dataset, namely Armeni~\etal \cite{armeni_iccv19} is listed in Tbl. \ref{table:dataset_comparision}. More information and statistics about \nameDataset{ }are provided in the supplementary. In the following a detailed description of the different entities of our 3D semantic scene graphs are given. 

\begin{table}[htbp]
\centering \def\arraystretch{1.1} \small
\caption{Semantic 3D scene graph comparision.}
\label{table:dataset_comparision}
\resizebox{\linewidth}{!}{%
{\renewcommand{\arraystretch}{1.3}%
\begin{tabular}{lllll}
dataset & size & instances & classes & obj. rel. \\
\hline
Armeni~\etal \cite{armeni_iccv19}* &
35 buildings & 3k & 28 & 4 \\
 & 727 rooms &  &  &  \\
\textbf{\nameDataset~(Ours)} & 1482 scans & \instanceCount & \labelCount & \relCount \\
 & 478 scenes & & & \\
\end{tabular}}}
\vspace{-0.3cm}
\end{table}

\subsection{Nodes}

\begin{figure}[ht!]
    \centering
    \includegraphics[width=0.95\linewidth]{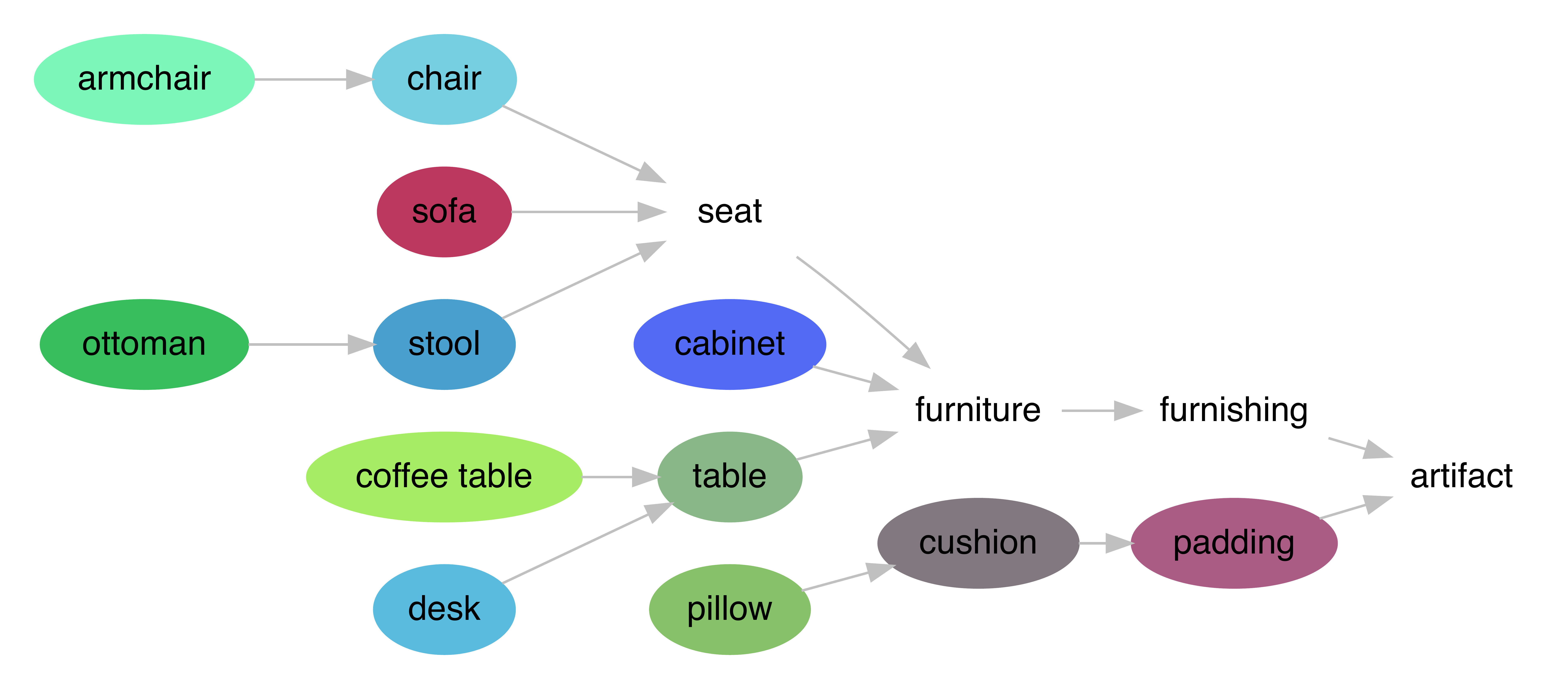}
    \caption{Simplified graphical visualization of the lexical relationships on a small subset of classes} 
    \label{fig:word_graph}
\end{figure}

The nodes in our graph are per definition 3D object instances, and each instance is assigned to a 3D scene. Each instance is defined by a class hierarchy where the class of order 1, $c_1$, in $c$ is the corresponding annotated label. The subsequent class labels are acquired by recursively parsing the lexical definition for hypernyms of $c_1$ using WordNet \cite{Fellbaum1998}. The definition \textit{``chair with a support on each side for arms"} gives us $c_{n+1} = \text{\rel{chair}}$ as a hypernym for $c_{n} = \text{\rel{armchair}}$. 
Lexical ambiguities result in multiple interpretations of a class label (lexeme); therefore a selection step is required to get only a single definition per class that is most likely in an indoor setting. Given the fact that the ~1.5k 3D reconstructions feature \labelCount different class labels, \labelCount lexical descriptions and their corresponding class hierarchy are provided. Fig. \ref{fig:word_graph} visualizes the lexical relationships on a small subset of classes. A more complete graph can be found in the supplementary. 

\subsection{Attributes}
\label{sec:attributes}

Attributes are semantic labels that describe object instances. 
This includes static and dynamic properties, as well as affordances.
Due to the large number of object instances and the desired semantic diversity of the attributes, an efficient extraction and annotation design is crucial. In the following we define the different types of attributes and their acquisition.

\paragraph{Static Properties} include visual object features such as the color, size, shape or texture but also physical properties \eg the (non-)rigidity. Geometric data and class labels are utilized to identify the relative size of the object in comparison with other objects of the same category. Since some features are class specific, we assign them on the class level. An example is an automatic attribute extraction from the lexical descriptions \eg a ball is \rel{spherical}. The remaining, more complex, attributes such as the material (\rel{wooden, metal}), shape (\rel{rectangular, L-shaped}) or the texture (color or pattern) are instance specific and manually annotated by expert annotators with an interface that was specifically designed for this purpose. We annotate static attributes in the reference scan and copy to each rescan, since they are not subject of change.

\paragraph{Dynamic Properties} are particularly important object attributes, which we refer to as states, such as \rel{open / closed}, \rel{full / empty} or \rel{on / off}. We define a state category to be class specific, while its current condition is a matter of instance and therefore also annotated with the aforementioned interface, together with generic static properties. Since state properties of objects can change over time specific instances in the rescans are separately annotated.

\paragraph{Affordances} Following previous works \cite{Gibson1979, xiazamirhe2018gibsonenv, armeni_iccv19} we define affordances as interaction possibilities or object functionalities of nodes of a specific object class \eg a \rel{seat} is for \rel{sitting}. We however condition them with their state attribute: only a \rel{closed door} can be \rel{opened}. This is particularly interesting since our 3D scans are from changing scenes. These changes often involve state changes caused by human interaction (see examples in the supplementary material). Overall, \nameDataset{ }features \numberAttributes{ }different attributes on approx. \numberAttributesInstances{ }object instances and \numAttributesTotal{ }attributes in total.

\subsection{Relationships}
\nameDataset{ }has a rich set of relationships classifiable into a) spatial / proximity relationships b) support relations and c) comparative relationships. 

\paragraph{Support Relationships}
Support relationships indicate the supporting structures of a scene \cite{Silberman2012}. By definition, an instance can have multiple supports; walls are by default supported by the floor and the floor is the only instance that, by definition, does not have any support. Automatically extracting support relationships is quite challenging due to the noisy and partial nature of real 3D scans. For each object in the scene, we consider neighbouring instances in a small radius (\eg 5cm) support candidates. These support candidates then undergo a verification procedure to a) eliminate incorrect supports and b) complete missing candidates. Remaining class-to-class (\eg \rel{bottle-table}) support pairs are then annotated with a so called \textit{semantic support} (\eg \rel{standing, lying}) and then specified for each instance in the dataset. 

\paragraph{Proximity Relationships}
Proximity relationships describe the spatial relationships (\eg \rel{next to, in front of}) with respect to a reference view. To limit redundancy, we only compute proximity relationships between the nodes that share a support parent. A bottle on a table therefore has no proximity relationship with a chair but the supporting table does, since the proximity relationship of the bottle can automatically be derived from its support parent.  

\begin{figure*}[ht]
\begin{center}
   \includegraphics[width=\linewidth]{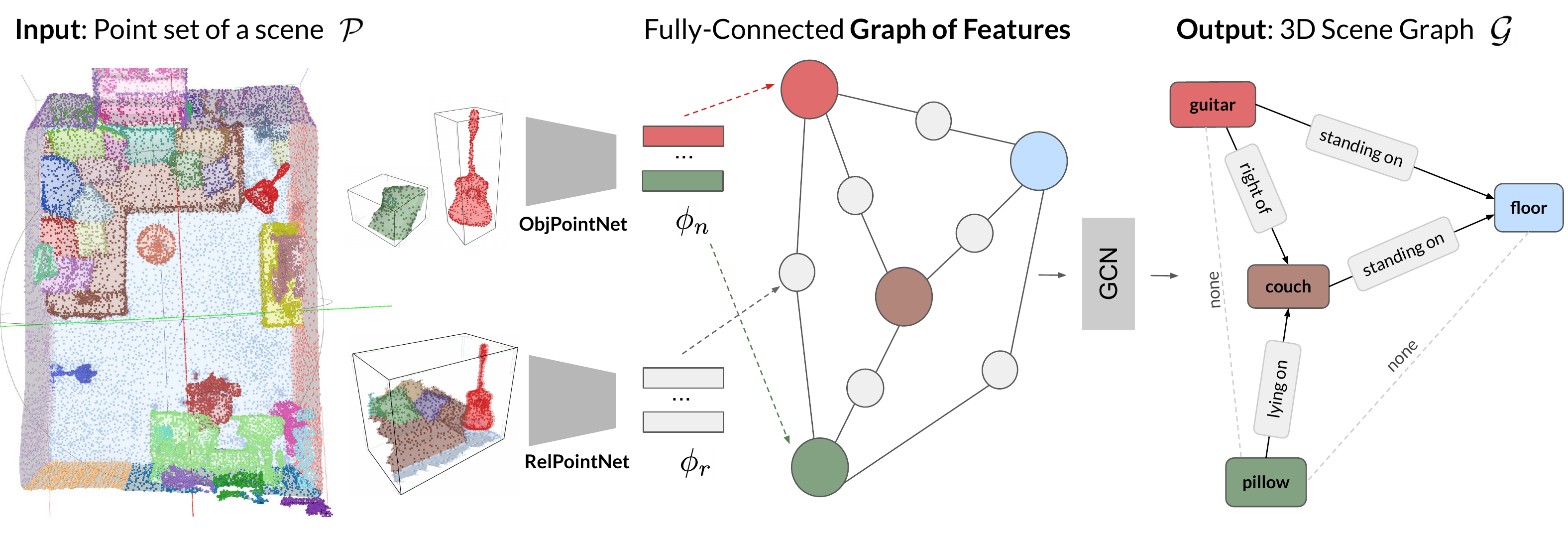}
\end{center}
   \caption{\textbf{Scene Graph Prediction Network} Given a point set $\mathcal{P}$ of a scene, annotated with instance segmentation $\mathcal{M}$, we infer a scene graph $\mathcal{G}$. \emph{Left:} Visual point features $\phi$ are extracted per object (color-coded) and per edge. \emph{Center:} The features $\phi$ are arranged in a graph structure for further processing from a GNN. \emph{Right:} The predicted graph, consisting of labeled object nodes and directed labeled edges.}
\label{fig:method_gp}
\end{figure*}

\paragraph{Comparative Relationships} The last group of relationships are derived from comparison of attributes, \eg \rel{bigger than, darker than, cleaner than, same shape as}. We use aforementioned attributes, see Section \ref{sec:attributes}, to generate these.
\section{Graph Prediction}

Given the point set $\mathcal{P}$ of a scene $s$ and the class-agnostic instance segmentation $\mathcal{M}$, the goal of the Scene Graph Prediction Network (SGPN) is to generate a graph $\mathcal{G} = (\mathcal{N}, \mathcal{R})$, describing the objects in the scene $\mathcal{N}$ as well as their relationships $\mathcal{R}$, Fig. \ref{fig:method_gp}. We base our learning method on a common principle in scene graph prediction \cite{Lu2016,xu2017scenegraph,yang2018graph}, which involves extraction of visual features for every node $\phi_n$ and edge $\phi_r$. We use two PointNet \cite{Ruizhongtai2016} architectures for the extraction of $\phi_n$ and $\phi_r$, which we dub namely \npnet and RelPointNet. For a scene $s$, we extract the point set of every instance $i$ separately, masked with $\mathcal{M}$
\begin{equation}
    \mathcal{P}_i = \{\delta_{m_k i}\odot p_k\}_{k=1,|\mathcal{P}|}
\end{equation}

\noindent where $\delta$ represents the Kronecker delta \footnote{$\delta_{ij}=1 \iff i=j$}, $p,m$ are instances of $\mathcal{P,M}$ and $|\cdot|$ is the cardinality of $\mathcal{P}$, \ie the number of points.
Each of the individual point sets $\mathcal{P}_i$ is the input to \npnet. 

Additionally, we extract a point set for every pair of nodes $i$ and $j$, using the union of the respective 3D bounding boxes $\mathcal{B}$

\begin{equation}
    \mathcal{P}_{ij} = \{p_k | p_k \in (\mathcal{B}^i \cup \mathcal{B}^j)\}_{k=1,|\mathcal{P}|}.
\end{equation}

\noindent The input to \epnet is a point set $\mathcal{P}_{ij}$, concatenated with the respective mask $\mathcal{M}_{ij}$, which is one if the point corresponds to object $i$, two if the point corresponds to object $j$ and zero otherwise. Preserving the orientation of the edge context $\mathcal{P}_{ij}$ is important to infer proximity relationships like \rel{left} or \rel{right}. Therefore, we disable rotational augmentation. We normalize the center of the object and edge point clouds, before feeding them to the respective networks. We arrange the extracted features in a graph structure, in the form of relationship triples (\rel{subject, predicate, object}), where $\phi_n$ occupy subject / object units, while edge features $\phi_r$ occupy the predicate units. 

We employ a Graph Convolutional Network (GCN) \cite{kipf2017semi}, similar to \cite{johnson2018image}, to process the acquired triples. As scenes come with diverse complexities, we want the GCN to allow flexibility in the number of input nodes. Each message-passing layer $l$ of the GCN consists of two steps. First, each triplet $ij$ is fed in an MLP $g_1(\cdot)$ for information propagation
\begin{equation}
    (\psi_{s,ij}^{(l)}, \phi_{p,ij}^{(l+1)}, \psi_{o,ij}^{(l)}) = g_1(\phi_{s,ij}^{(l)}, \phi_{p,ij}^{(l)}, \phi_{s,ij}^{(l)})
\end{equation}

\noindent where $\psi$ represent the processed features, $s$ indicates subject, $o$ indicates object, and $p$ predicate. Second, for a certain node, in an aggregation step, the signals coming from all the valid connections of that node (either as a subject or an object) are averaged together
\begin{equation}
    \rho_i^{(l)} = \frac{1}{|\mathcal{R}_{i,s}| + |\mathcal{R}_{i,o}|} \Big(\sum_{j\in \mathcal{R}_s} \psi_{s,ij}^{(l)} + \sum_{j\in \mathcal{R}_o} \psi_{o,ji}^{(l)}\Big)
\end{equation}
\noindent where $|\cdot|$ denotes cardinality and $\mathcal{R}_s$ and $\mathcal{R}_o$ are the set of connections of the node as subject and as objects respectively. The resulting node feature is fed in another MLP $g_2(\cdot)$. Inspired by \cite{li2019deepgcns}, we adapt a residual connection to overcome potential Laplacian smoothing on graphs and obtain the final node feature as
\begin{equation}
    \phi_i^{(l+1)} = \phi_i^{(l)} + g_2(\rho_i^{(l)}).
\end{equation}

\noindent The final features $\phi_{s,ij}^{(l+1)}, \phi_{p,ij}^{(l+1)}, \phi_{o,ij}^{(l+1)}$ are then processed by the next convolutional layer $l$, in the same fashion. After each layer $l$, the node visibility is propagated to a further neighbour level. Hence, the number of layers equals the order of relations that the model can capture.

The last part of the GCN consists of two MLPs for the prediction of the node and predicate classes. 

\paragraph{Losses} We train our model end-to-end, optimizing an object classification loss $\mathcal{L}_{\rm{obj}}$ as well as a predicate classification loss $\mathcal{L}_{\rm{pred}}$

\begin{equation}
\mathcal{L}_{\rm{total}} = \lambda_{obj} \mathcal{L}_{\rm{obj}} + \mathcal{L}_{\rm{pred}}
\end{equation}

\noindent where $\lambda_{obj}$ is a weighting factor. We assume that, realistically, for a certain object pair there are multiple valid relationships that describe their interaction. For instance, in Fig. \ref{fig:teaser}, a chair can be \rel{front of} another chair, while simultaneously having the same appearance (\rel{same as}). Therefore, we formulate $\mathcal{L}_{\rm{pred}}$ as per-class binary cross entropy. This way, it is judged independently whether an edge should be assigned a certain label (\eg \rel{standing on}) or \rel{none}. To deal with class imbalance, for both loss terms we use a focal loss \cite{Lin2017FocalLoss}
\begin{equation}
    \mathcal{L} = - \alpha_t (1 - p_t) ^ \gamma \log{p_t}
\end{equation}

\noindent where $p_t$ represents the logits of a prediction and $\gamma$ is a hyper-parameter. $\alpha_t$ is the normalized inverse frequency for the multi-class loss ($\mathcal{L}_{obj}$) and a fixed edge / no-edge factor for the per-class loss ($\mathcal{L}_{pred}$). 

\smallskip
Implementation details are provided in the supplement.  
\section{Scene Retrieval}

\begin{figure}[th!]
\begin{center}
   \includegraphics[width=\linewidth]{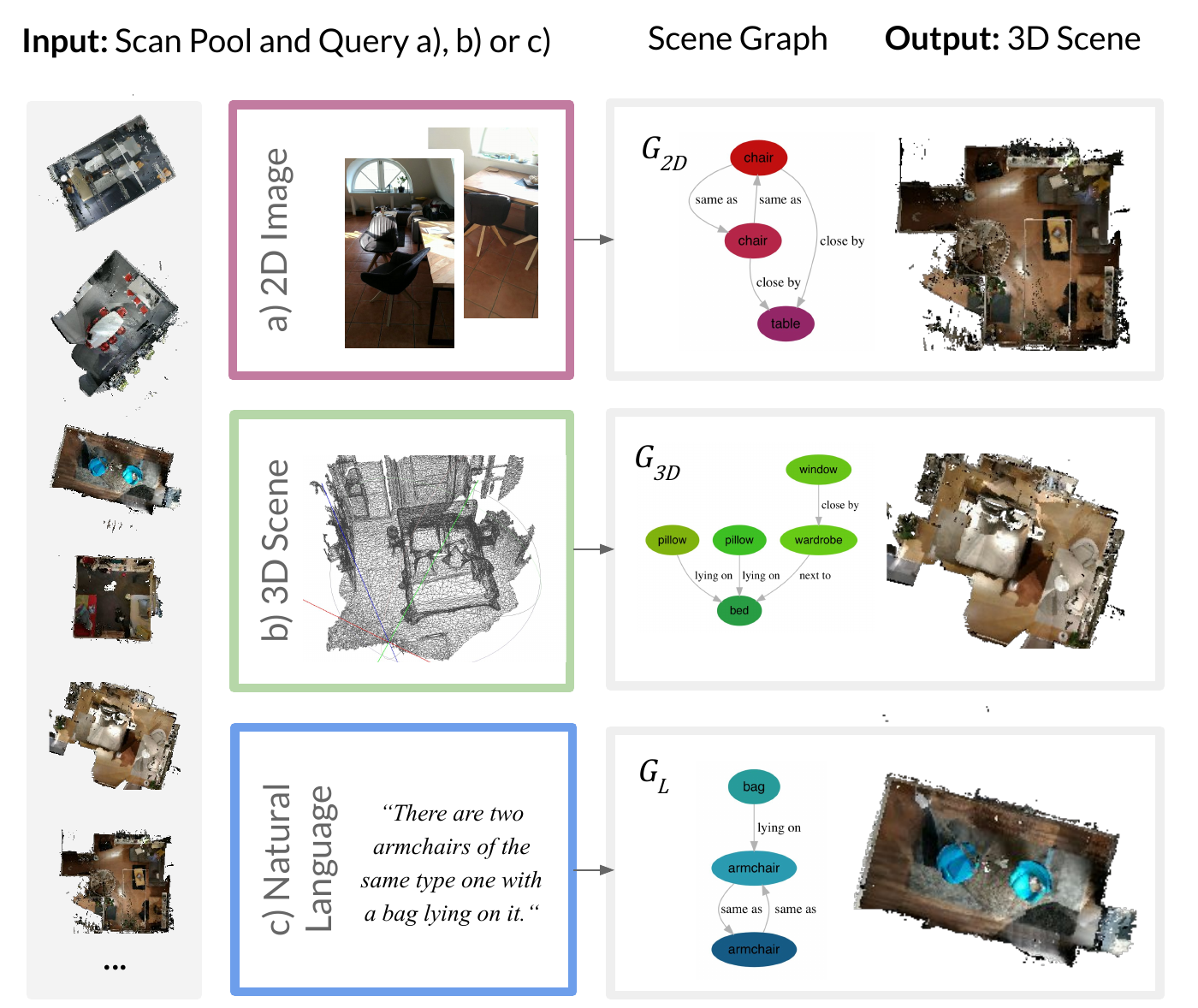}
\end{center}
   \caption{\textbf{Cross Domain 2D-3D Scene Retrieval:} scene graphs are used in our cross-domain scene retrieval task to close the domain gap between 2D images, 3D scenes and other modalities}
\label{fig:method_r}
\end{figure}

\begin{figure*}[h!]
    \centering
    \includegraphics[width=\linewidth]{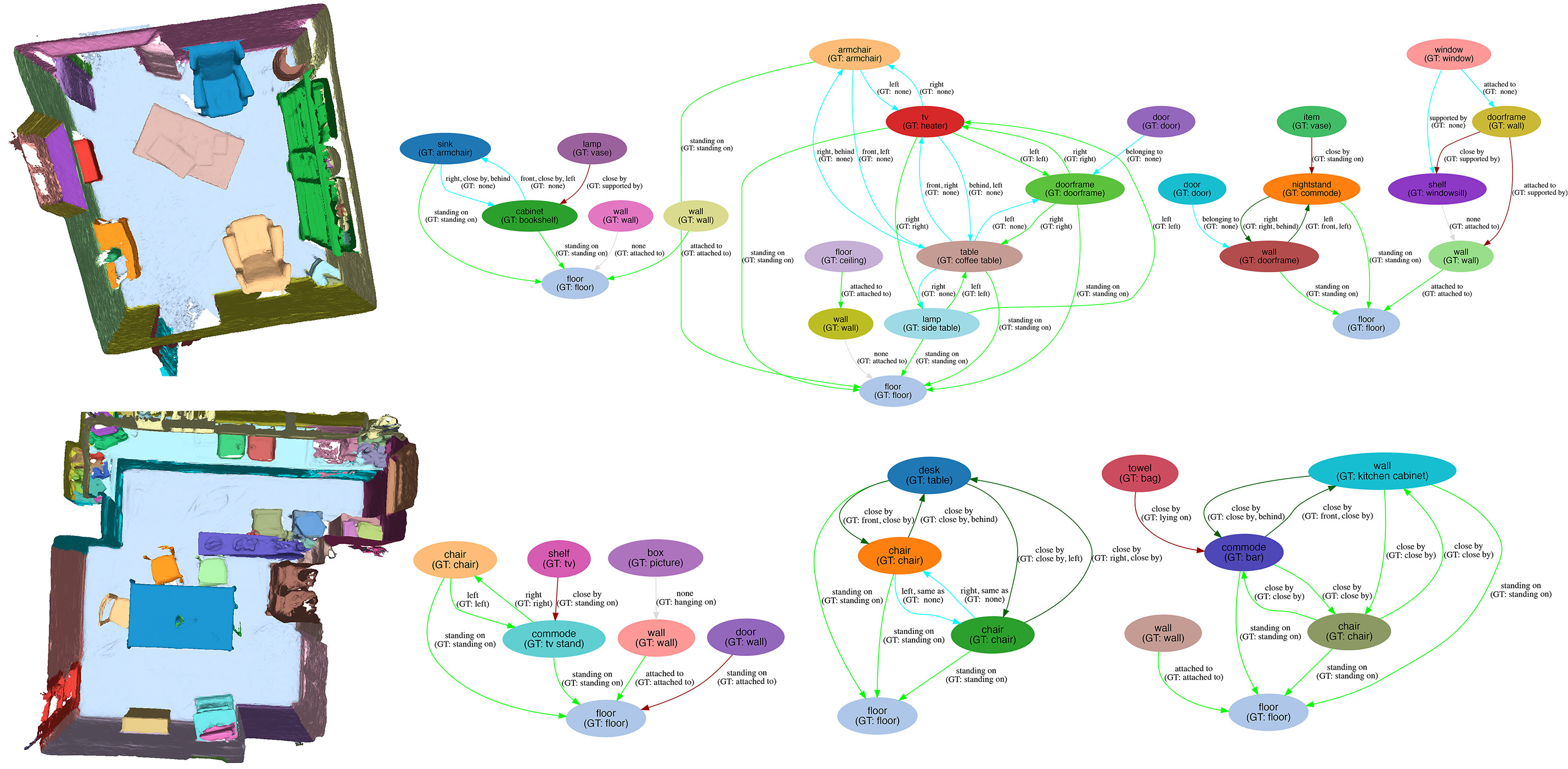}
    \caption{\textbf{Qualitative results of our scene graph prediction model} (best viewed in the digital file). \emph{Green}: correctly predicted edges,
    \emph{blue}: missing ground truth, \emph{red}: miss-classified edges, \emph{gray}: wrongly predicted as \emph{none} when GT is a valid relationship.}
    \label{fig:graph_pred}
\end{figure*}

We introduce a new cross-domain task named \textit{image based 3D scene retrieval in changing indoor environments} which is about identifying the 3D scene from a list of scans given a single 2D image with potential global and local changes (see Fig. \ref{fig:method_r}). This is particularly challenging since it involves a) multiple domains (2D images and 3D models) and b) scene changes (moving objects, changing illumination). For the evaluation we select semantically rich 2D images from the rescan sequences in 3RScan \cite{Wald2019RIO}. Due to the domain gap between 2D images and 3D we propose carrying out this novel retrieval task through scene graphs -- which are by definition more stable towards scene changes -- and serve as a shared domain between 2D and 3D. Such an approach also allows to retrieve 3D scenes from any input domain from which a scene graph can be generated \eg natural language or 3D directly. We show how different similarity metrics can be used to successfully find the correct 3D scene using not only object semantics but also the scene context in form of semantically meaningful relationships between object instances. Computing the similarity between graphs is a NP-complete problem, so instead of matching graphs directly via their graph edit distance we first transform our scene graphs into multisets containing node classes and their (semantic) edges / tuples. Please note that these potentially have repetition of elements. To get the similarity of of two graphs, a similarity scores $\tau$ is applied on the corresponding multisets $s(\mathcal{G})$ respectively. For our tests we explore two different similarity functions: Jaccard $\tau_J(A,B)$, eq. \ref{eq:Jaccard} and Szymkiewicz-Simpson $\tau_S(A,B)$, eq. \ref{eq:SzymkiewiczSimpson}. 

\begin{equation}
\label{eq:Jaccard}
    \tau_J(A,B) = \frac{|A \cap B|}{|A \cup B|}
\end{equation}

\begin{equation}
\label{eq:SzymkiewiczSimpson}
    \tau_S(A,B) = \frac{|A \cap B|}{\min(|A|, |B|)}
\end{equation}

While the Jaccard coefficient is a widely used metric, the Szymkiewicz-Simpson coefficient can provide more meaningful similarity scores especially when the two sets $A$ and $B$ have very different sizes which is often the case in a 2D-3D scenario. When matching two graphs $\mathcal{G}$ and $\mathcal{G}'$ we combine the similarity metric of the object semantics, generic node edges $\mathcal{E}$ as well as semantic relationships $\mathcal{R}$ and obtain

\begin{equation}
\label{eq:retrieval_eq}
    f(\hat{\mathcal{G}}, \hat{\mathcal{G}}')= \frac{1}{|\hat{G}|} \sum_{i = 1}^{|\hat{\mathcal{G}}|} \tau(s(\hat{\mathcal{G}}^{(i)}), s(\hat{\mathcal{G}}'^{(i)}))\footnote{we define $f_S$ and $f_J$ to use $\tau_S$ and $\tau_J$ respectively.}
\end{equation}

where $\tau$ is either the Jaccard or Szymkiewicz-Simpson coefficient and $\hat{\mathcal{G}}$ is defined as the augmented graph $\hat{\mathcal{G}} = (\mathcal{N}, \mathcal{E}, \mathcal{R})$ where $\mathcal{E}$ are binary edges. Interestingly, one can use our retrieval method to find rooms that fullfill certain requirements such as the available of objects \eg meeting room with a \rel{TV, whiteboard} but could also include affordances: \rel{sitting} for 20 people.
\section{Evaluation}

\begin{table*}[htbp]
\caption{Evaluation of the scene graph prediction task on \nameDataset{}. We present triples prediction, object classification as well as predicate prediction accuracy.}
\label{table:predicate_prediction}
{\renewcommand{\arraystretch}{1.3}%
\begin{tabular}{lcccccc}
       & \multicolumn{2}{c}{Relationship Prediction} & \multicolumn{2}{c}{Object Class Prediction} & \multicolumn{2}{c}{Predicate Prediction}   \\
Method & R@50       & R@100       & R@5       & R@10       & R@3       & R@5   \\
\hline
\ding{192} Relation Prediction Baseline  &     0.39      &      0.45    &    0.66        &     0.77        &       0.62      &     0.88     \\
Single Predicate, ObjCls from PointNet Features & 0.37 & 0.43 & \textbf{0.68} & \textbf{0.78} & 0.42 & 0.58\\
\ding{193} Multi Predicate, ObjCls from PointNet Features &    \textbf{0.40}        & \textbf{0.66}            &     \textbf{0.68}      &    \textbf{0.78}         &     \textbf{0.89}        &    \textbf{0.93}          \\
Multi Predicate, ObjCls from GCN Features  &      0.30     &   0.60       &      0.60      &    0.73         &      0.79      &      0.91        \\
\end{tabular}}
\end{table*}

In the following, we first report results of our 3D graph prediction by comparing it against a relationship prediction baseline, inspired by \cite{Lu2016}, on our newly created \nameDataset-dataset. We re-implemented and adapted their method to work with 3D data. The baseline extracts node and edge features from an image, which we translate to PointNet features in 3D, similar to our network. The edge and node features are passed directly, namely to a predicate and object classifier. For evaluation we use the same train and test splits as originally proposed by \cite{Wald2019RIO}. We validate the effectiveness of our multi predicate classifier and GCN in our proposed network in an ablation study. In the second section, we evaluate different graph matching functions in 2D-3D as well as 3D-3D retrieval by matching changed scenes.

\subsection{Semantic Scene Graph Prediction}

Here, we report the results of our scene graph prediction task. Following previous works \cite{xu2017scenegraph} we first separately evaluate the predicate (relationship) prediction in isolation from the object classes. The overall scene graph prediction performance is evaluated jointly where the relationship as well as the object categories are to be predicted given a set of localized objects. Since our method predicts the relationship as well as the object categories independently from another, we obtain an ordered list of triplet classification scores by multiplying the respective scores \cite{yang2018graph}. Similarly to the predicate prediction, the performance of the object categories is reported. We adopt the recall metric used in \cite{Lu2016} to evaluate most confident (\texttt{subject, predicate, object}) triplets against the ground-truth in a top-n manner. Tbl. \ref{table:predicate_prediction} shows that we outperform the baseline in graph related metrics, while being on par in object classification. Additionally, as expected, the multiple predicate prediction model leads to a higher predicate accuracy, which we attribute to the inherent ambiguity in a single classification problem, when multiple outputs are plausible. Moreover, we compare two versions of our model, in which the object classification is performed a) directly on the PointNet features $\phi_n$ and b) to the output of the GCN. We observe a slight improvement in the object and predicate accuracy for the former. Fig. \ref{fig:graph_pred} illustrates the predicted scene graphs. In all edges and nodes we show the predictions together with the ground truth in brackets. More examples can be found in the supplement.

\subsection{Scene Retrieval}

Tbl. \ref{table:scene_retrieval3D} and \ref{table:scene_retrieval2D} report two scene retrieval tasks.\footnote{In the tables we replace $\hat{\mathcal{G}}$ with $\mathcal{G}$ for notation simplicity} The goal is to match either a single 2D image (Tbl. \ref{table:scene_retrieval2D}) or a 3D rescan of an indoor scene (Tbl. \ref{table:scene_retrieval3D}) with the most similar instance from a pool of 3D reference scans from the validation set of 3RScan. We compute the scene graph similarity between each rescan (2D or 3D) and the target reference scans. We then order the matches by their similarity and report the top-n metric, \ie the percentage of the true positive assignments, placed in the top-n matches from our algorithm. In our experiment, we either use ground truth or predictions for the query and source graphs (see Graph-column in Tbl. \ref{table:scene_retrieval3D} and \ref{table:scene_retrieval2D}). To measure the effect of the different similarity functions, decoupled from the graph prediction accuracy, we first evaluate $\tau_J(A, B)$ and $\tau_S(A, B)$ using ground truth graphs. Since the size of image and 3D scene graphs are significantly different, using the Szymkiewicz-Simpson coefficient in 2D-3D matching leads to better results while the performance of the Jaccard coefficient is on par in the 3D-3D scenario. We observe that adding semantic relationships to the graph matching improves the scene retrieval. The results also confirm that our predicted graphs \ding{193} achieve higher matching accuracy compared to the baseline model \ding{192}. Note that for the purpose of this experiment, predicted 2D graphs are obtained by rendering the predicted 3D graphs as described in Section \ref{sec:data}.

\begin{table}[htbp]
\centering
\caption{Evaluation: 3D-3D scene retrieval of changing 3D rescans to reference 3D scans in 3RScan.}
\label{table:scene_retrieval3D}
\resizebox{\linewidth}{!}{%
{\renewcommand{\arraystretch}{1.3}%
\begin{tabular}{lcllll}
 & Graph & Top-1 & Top-3 & Top-5 \\
\hline
$\tau_S(s(\mathcal{N}_{3D}), s(\mathcal{N}_{3D}))$	&	GT	&0.86&	0.99&	1.00\\
$f_S(\mathcal{G}_{3D}, \mathcal{G}_{3D})$	&	GT	&0.96&	1.00&	1.00\\
$\tau_J(s(\mathcal{N}_{3D}), s(\mathcal{N}_{3D}))$	&	GT	&0.89&	0.95&	0.95\\
$f_J(\mathcal{G}_{3D}, \mathcal{G}_{3D})$	&	GT	&0.95&	0.96&	0.98\\
$\tau_J(s(\mathcal{N}_{3D}), s(\mathcal{N}_{3D}))$	& \ding{192} & 0.15&	0.40&	0.45\\
$f_J(\mathcal{G}_{3D}, \mathcal{G}_{3D})$	&\ding{192}	& 0.29&	0.50&	0.59\\
$\tau_J(s(\mathcal{N}_{3D}), s(\mathcal{N}_{3D}))$	&	\ding{193}	&	0.32&	0.46&	0.50\\
$f_J(\mathcal{G}_{3D}, \mathcal{G}_{3D})$	&	\ding{193}	& 0.34&	0.51&	0.56\\
\end{tabular}}}
\vspace{-0.3cm}
\end{table}

\begin{table}[htbp]
\centering
\caption{Evaluation: 2D-3D scene retrieval of changing rescans to reference 3D scans in 3RScan.}
\label{table:scene_retrieval2D}
\resizebox{\linewidth}{!}{%
{\renewcommand{\arraystretch}{1.3}%
\begin{tabular}{lclll}
 & Graph & Top-1 & Top-3 & Top-5 \\
\hline
$\tau_J(s(\mathcal{N}_{2D}), s(\mathcal{N}_{3D}))$	&	GT	&	0.49&	0.75&	0.84\\ 
$\tau_S(s(\mathcal{N}_{2D}), s(\mathcal{N}_{3D}))$	&	GT	&0.98&	0.99&	1.00\\
$f_J(\mathcal{G}_{2D}, \mathcal{G}_{3D})$	&	GT	& 0.55&	0.85&	0.86\\
$f_S(\mathcal{G}_{2D}, \mathcal{G}_{3D})$	&	GT	&	1.00&	1.00&	1.00\\
$\tau_S(s(\mathcal{N}_{2D}), s(\mathcal{N}_{3D}))$	& \ding{192} & 0.17&	0.36&	0.42\\
$f_S(\mathcal{G}_{2D}, \mathcal{G}_{3D})$	& \ding{192}	&0.10&	0.25&	0.32\\
$\tau_S(s(\mathcal{N}_{2D}), s(\mathcal{N}_{3D}))$	&	\ding{193}	&	0.17&	0.36&	0.41\\
$f_S(\mathcal{G}_{2D}, \mathcal{G}_{3D})$	&	\ding{193}	&	0.13&	0.38&	0.42\\
\end{tabular}}}
\vspace{-0.3cm}
\end{table}

\section{Conclusion}

In this work, we explore 3D semantic scene graphs. We release \nameDataset{, }a 3D scene graph dataset with semantically rich relationships based on 3RScan \cite{Wald2019RIO}. We use our data to train a graph prediction network for 3D scenes that is able to estimate not only object semantics but also relationships between objects. Further, we show the usefulness of graphs in 3D scenes by applying it to a new cross-domain task called \textit{image based 3D scene retrieval in changing indoor environments}. This shows how semantic scene graphs are useful to bridge the domain gap between 2D-3D; opening doors for new applications such as text-3D scene retrieval or VQA. We further believe that scene graphs (and their changes) could potentially help to better reason about human activities in changing indoor environments.

\section*{Acknowledgment}

We would like to thank Mariia Gladkova, Alina Karimova and Premankur Banerjee for their help with the data preparation and annotations. This work was funded by the Deutsche Forschungsgemeinschaft (DFG) $\#381855581$, the Bavarian State Ministry of Education, Science and the Arts in the framework of the Centre Digitisation Bavaria (ZD.B) and a Google AR/VR University Research Award. 

\newpage

\newpage

{\small
\bibliographystyle{ieee_fullname}
\bibliography{egbib}
}
\section{Supplementary Material}

\noindent
This document supplements our paper with additional details, visualization and results. Section \ref{sec:data_sup} contains more thorough statistics and characteristics regarding the \nameDataset{} dataset such as a visualization of the WordNet hierarchy, 2D graphs as well as the annotation interfaces. Section \ref{sec:graph_pred_sup} gives additional information about the proposed method as well as more graph prediction results while Section \ref{sec:retrieval} focuses on retrieval.

\subsection{\nameDataset{ }Dataset}
\label{sec:data_sup}

\begin{figure}[h!]
\begin{center}
    \includegraphics[width=\linewidth]{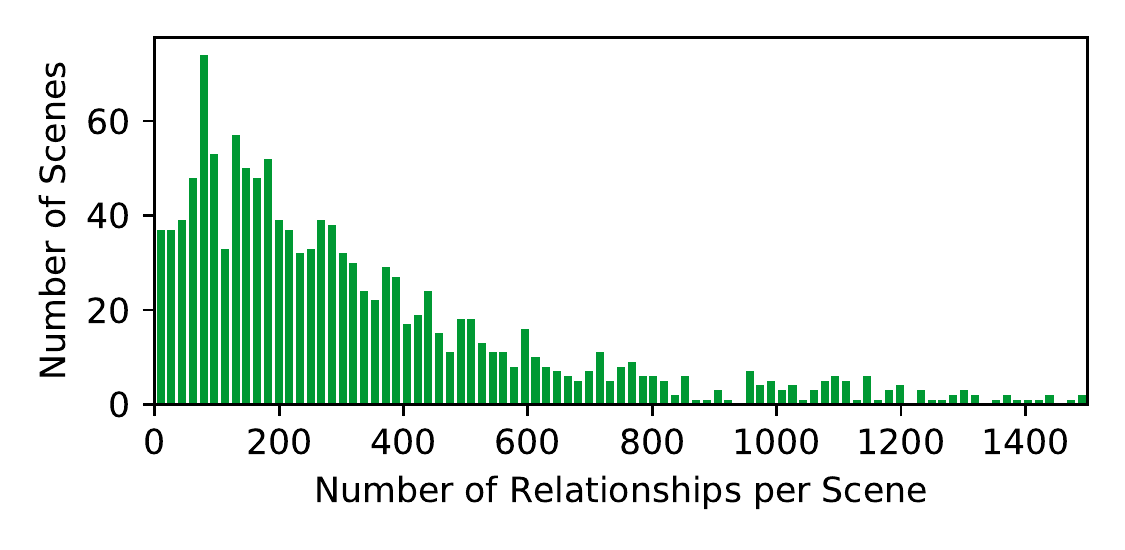}
\end{center}
   \caption{Histogram of scenes in \nameDataset{} and corresponding number of relationships}
    \label{fig:stats_scene_relationships}
\end{figure}

\begin{figure}[h!]
\begin{center}
    \includegraphics[width=\linewidth]{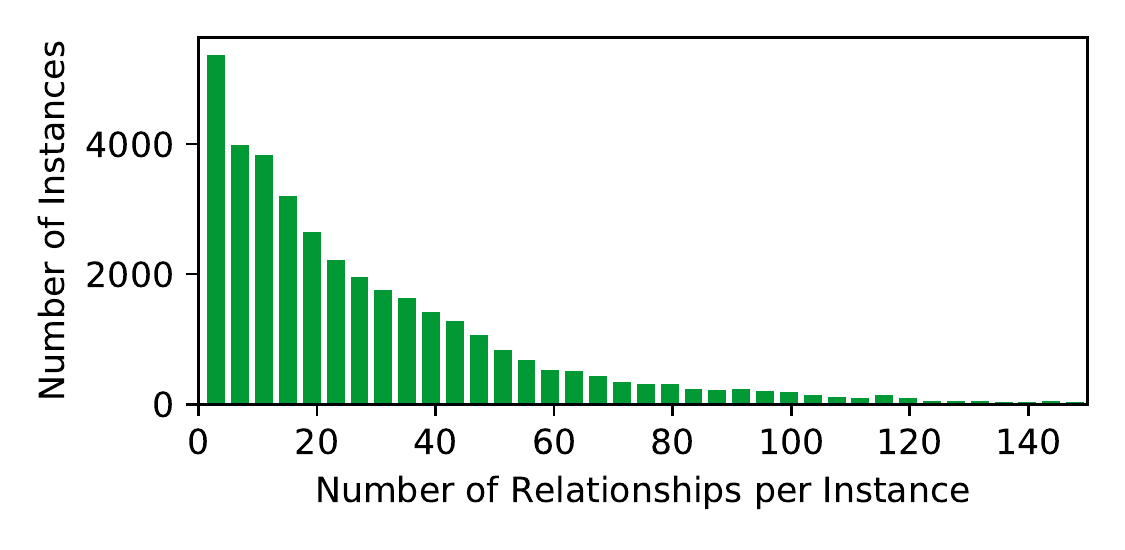}
\end{center}
   \caption{Histogram of object instances in \nameDataset{} and corresponding number of relationships}    \label{fig:relationships_objects_extended}
\end{figure}

\begin{figure}[h!]
\begin{center}
    \includegraphics[width=\linewidth]{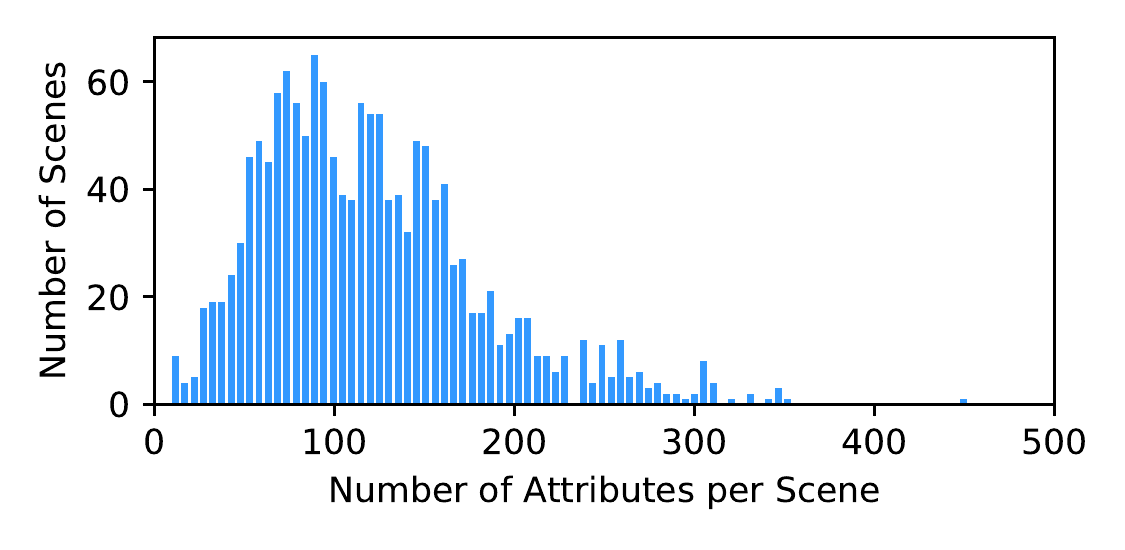}
\end{center}
   \caption{Histogram of scenes in \nameDataset{} and corresponding number of attributes}    \label{fig:stats_scene_attributes}
\end{figure}

\begin{figure}[h!]
\begin{center}
    \includegraphics[width=\linewidth]{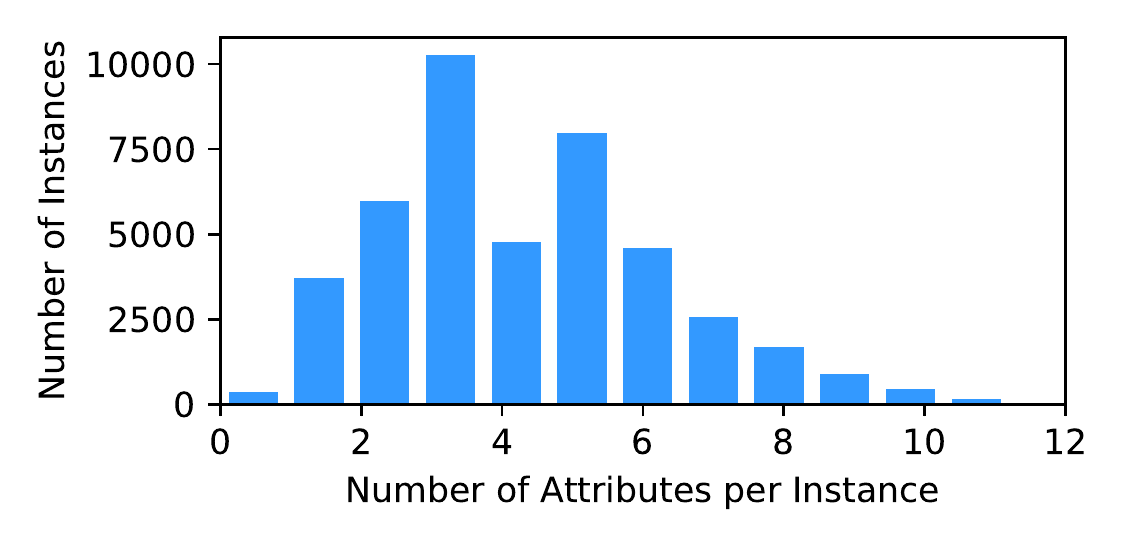}
\end{center}
   \caption{Histogram of object instances in \nameDataset{} and corresponding number of attributes}   \label{fig:stats_objects_attributes}
\end{figure}

\begin{figure*}[h]
\begin{center}
   \includegraphics[width=\linewidth]{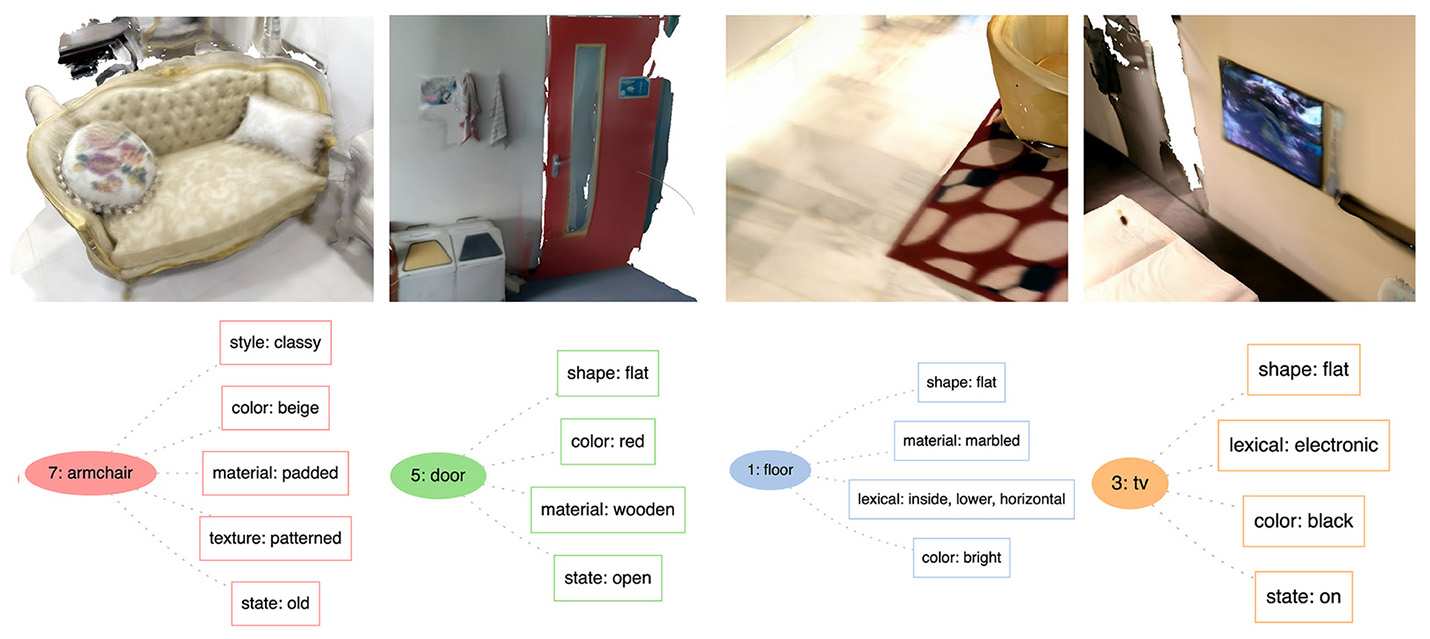}
\end{center}
   \caption{Example object instances (top) and their corresponding attributes (below).}
\label{fig:attribute_instances}
\end{figure*}

\paragraph{Statistics} In this paragraph, we present further data statistics. Fig. \ref{fig:stats_scene_relationships} and~\ref{fig:relationships_objects_extended} show the number of relationships per 3D scan and object instance. The corresponding histograms for object attributes are in Fig. \ref{fig:stats_scene_attributes} and \ref{fig:stats_objects_attributes}. Fig. \ref{fig:statistics_obj}, \ref{fig:stats_relationships}, \ref{fig:stats_attributes}, \ref{fig:stats_affordances} show the most frequent object, predicate, attribute and affordance occurrences extracted from our ground truth graphs. Fig. \ref{fig:semantic_graph_rio} highlight some of the most common semantic connections present in the dataset. Further, a few example of object instances and the annotated attributes can be found in Fig. \ref{fig:attribute_instances}. These statistics show that the scene graphs in \nameDataset{ }are not only semantically rich but also very dense.

\paragraph{WordNet Graphs} Fig. \ref{fig:word_rel_graph} shows a graphical visualization of the WordNet hierarchy of classes, which we use to extract the  per-node hierarchy of labels $c$. Colored nodes show class labels from the annotation set, while white nodes are abstract representations that are not part of the original label set. For an instance annotated as \rel{chair}, the hierarchical label $c$ would be $c = \{\rel{chair, seat, furniture, ..., entity}\}$. 

\paragraph{2D Graphs: Depth and Mask} Fig. \ref{fig:graph2D} illustrates 2D scene graphs of the \nameDataset{ }dataset, which are obtained via rendering the 3D scene. We show that, while 2D scene graph datasets currently available \cite{krishna2017visual} only have bounding box annotations, we also provide depth and dense semantic instance masks, which we believe are relevant for the future, to explore alternative ways for the graph prediction and other underlying tasks. 

\begin{figure*}[h]
\begin{center}
   \includegraphics[width=\linewidth]{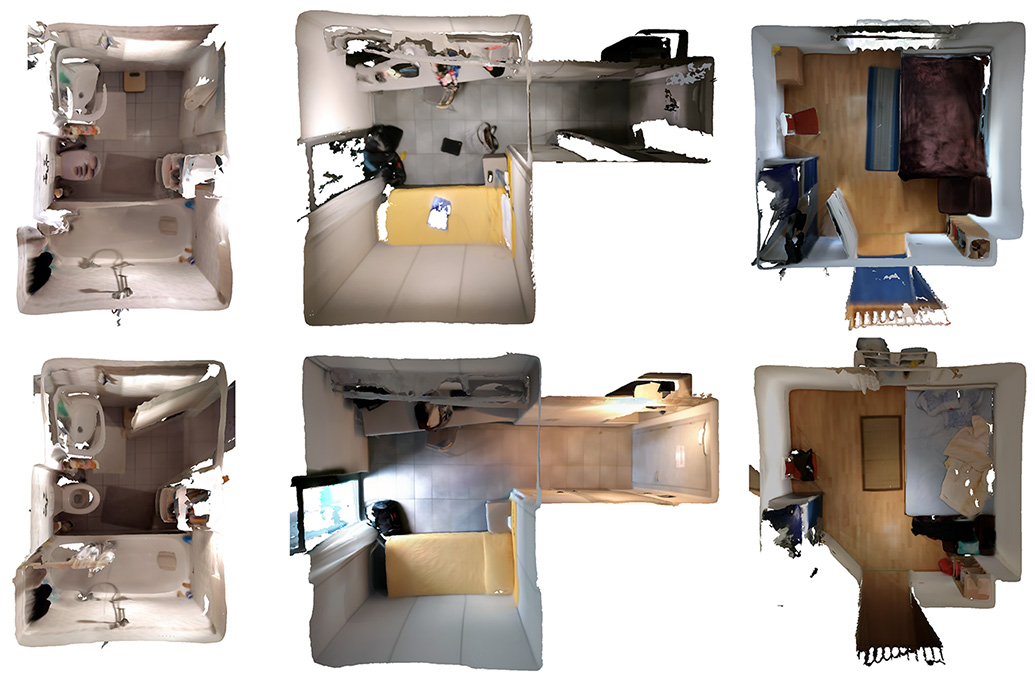}
\end{center}
   \caption{3 example scenes at two different time steps where human activity possibly have changed object states. \textit{Left:} someone might have used the toilet (toilet seat is \rel{down / up}), \textit{Center:} someone might has cleaned this room (\rel{desk} and \rel{floor} are \rel{messy / tidy}), 
   \textit{Right:} someone might have slept in the bed (\rel{bed} is \rel{tidy / messy}).}
\label{fig:state_changes}
\end{figure*}

\paragraph{State Changes} While static instance attributes such as the color or material of an object do not change, dynamic instance attributes (\eg \rel{on / off, open / closed}) can change over time. Interestingly, these state properties are closely connected to human interaction and could potentially give information about activities that might have happened in a particular 3D space (see Fig. \ref{fig:state_changes}).

\begin{figure*}[t]
\begin{center}
   \includegraphics[width=0.85\linewidth]{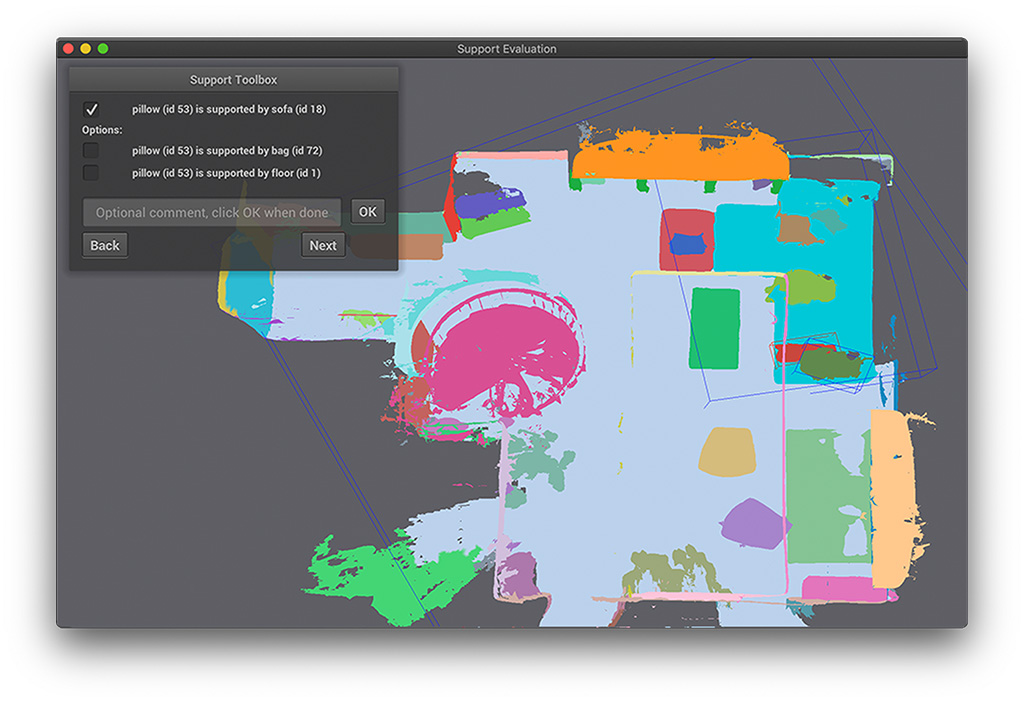}
\end{center}
   \caption{Interface for binary support annotation}
\label{fig:interface_binary_rel}
\end{figure*}

\begin{figure*}[t]
\begin{center}
   \includegraphics[width=0.85\linewidth]{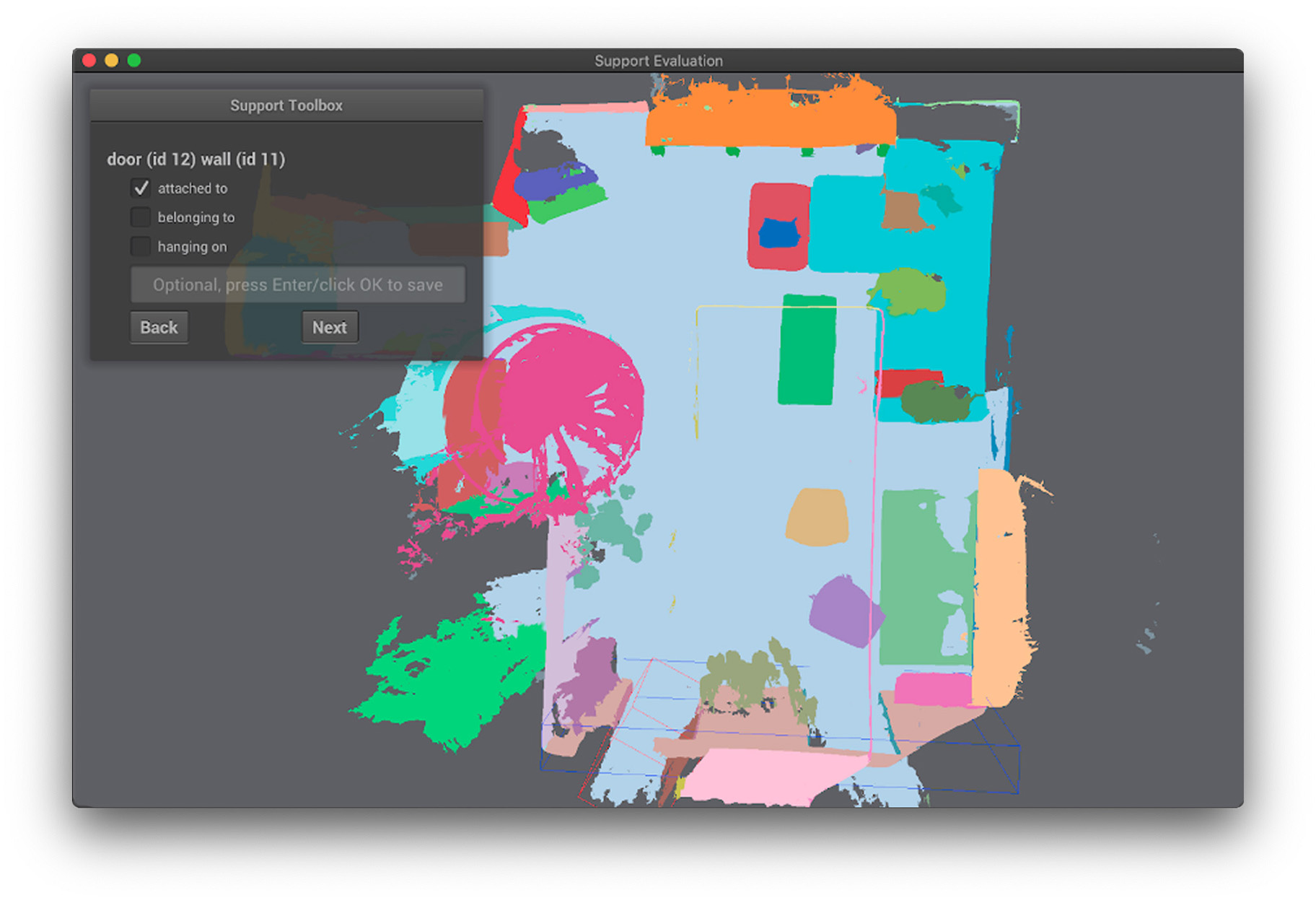}
\end{center}
   \caption{Interface for semantic relationship annotation}
\label{fig:interface_semantic_rel}
\end{figure*}

\begin{figure*}[t]
\begin{center}
   \includegraphics[width=\linewidth]{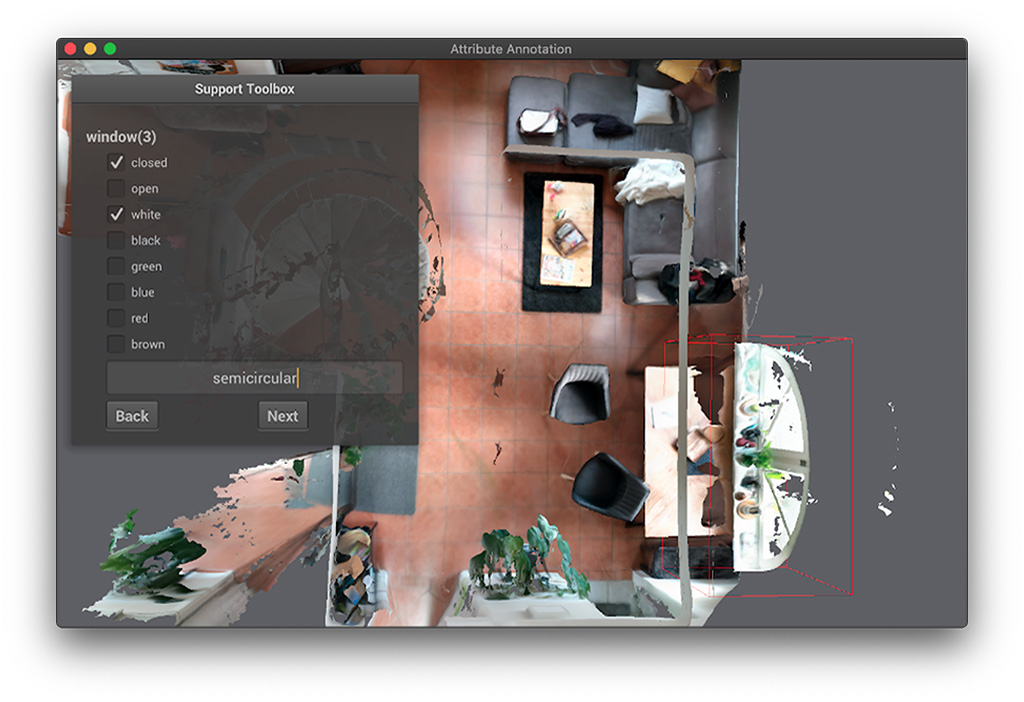}
\end{center}
   \caption{Annotation Interface for attribute annotation}
\label{fig:interface_atttributes}
\end{figure*}

\paragraph{Annotation interfaces} Fig. \ref{fig:interface_binary_rel}, \ref{fig:interface_semantic_rel} and \ref{fig:interface_atttributes} are screenshots of the user interfaces we used for the annotation, namely attribute and support relationships (for binary and semantic annotation and verification). Please note that semantic support annotations are done after the binary annotation, since they build upon the ground truth support pairs. 

\begin{figure}[t]
\begin{center}
    \includegraphics[height=0.9\textheight]{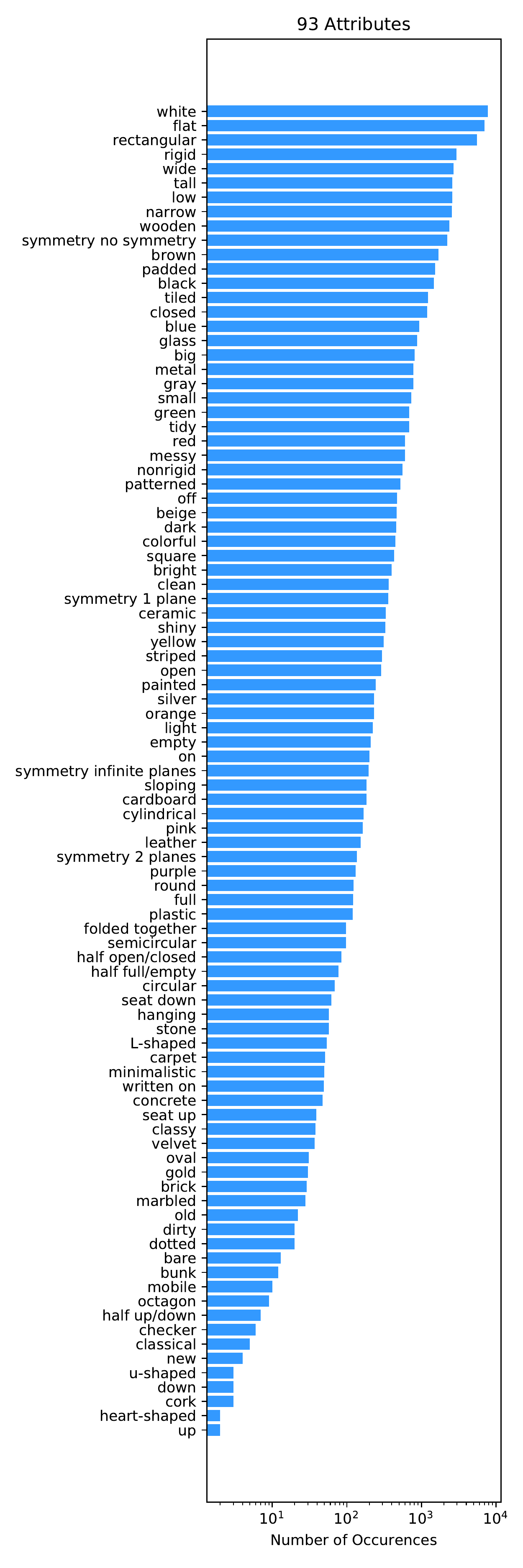}
\end{center}
   \caption{Attributes in the \nameDataset{} dataset, sorted by occurrence, presented in logarithmic scale}
    \label{fig:stats_attributes}
\end{figure}

\begin{figure}[t]
\begin{center}
    \includegraphics[height=0.9\textheight]{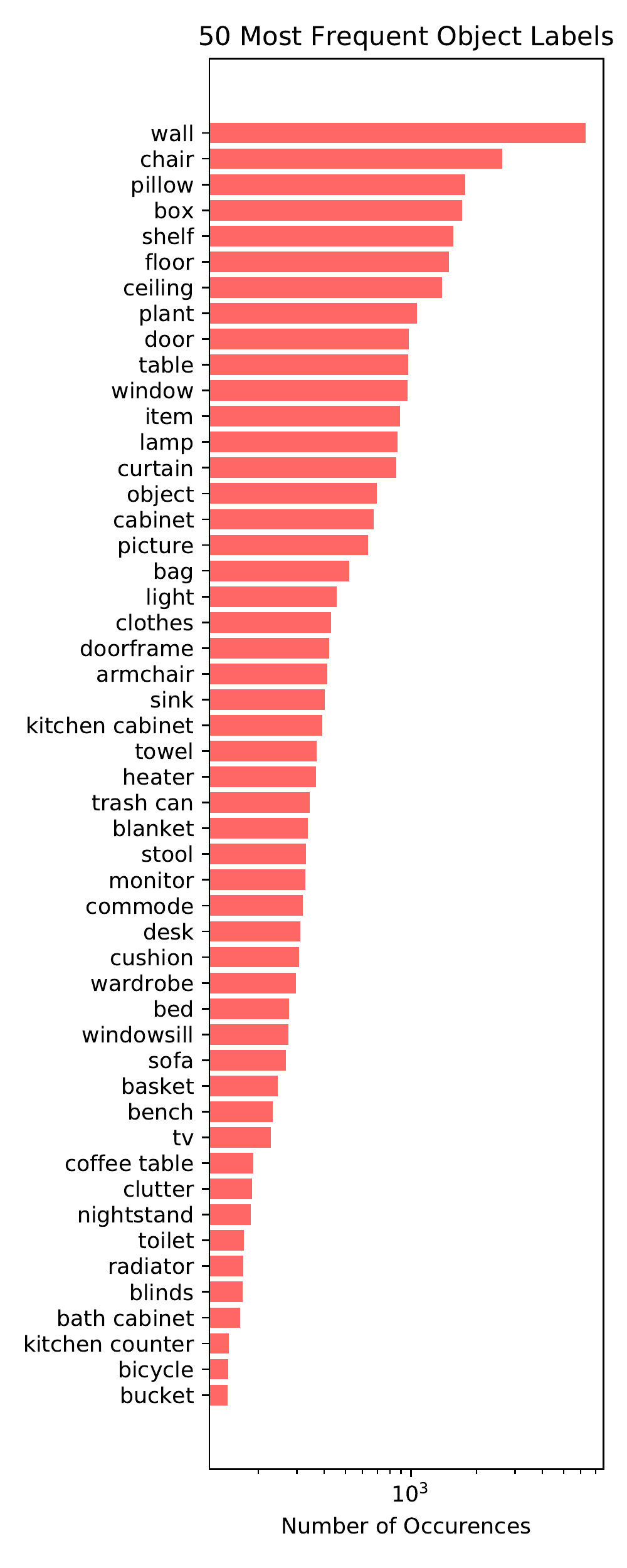}
\end{center}
   \caption{Most frequent (Top-50) object classes used for the training of the Scene Graph Prediction Network, sorted by occurrence, presented in logarithmic scale}
\label{fig:statistics_obj}
\end{figure}

\begin{figure}[h]
\begin{center}
    \includegraphics[height=0.9\textheight]{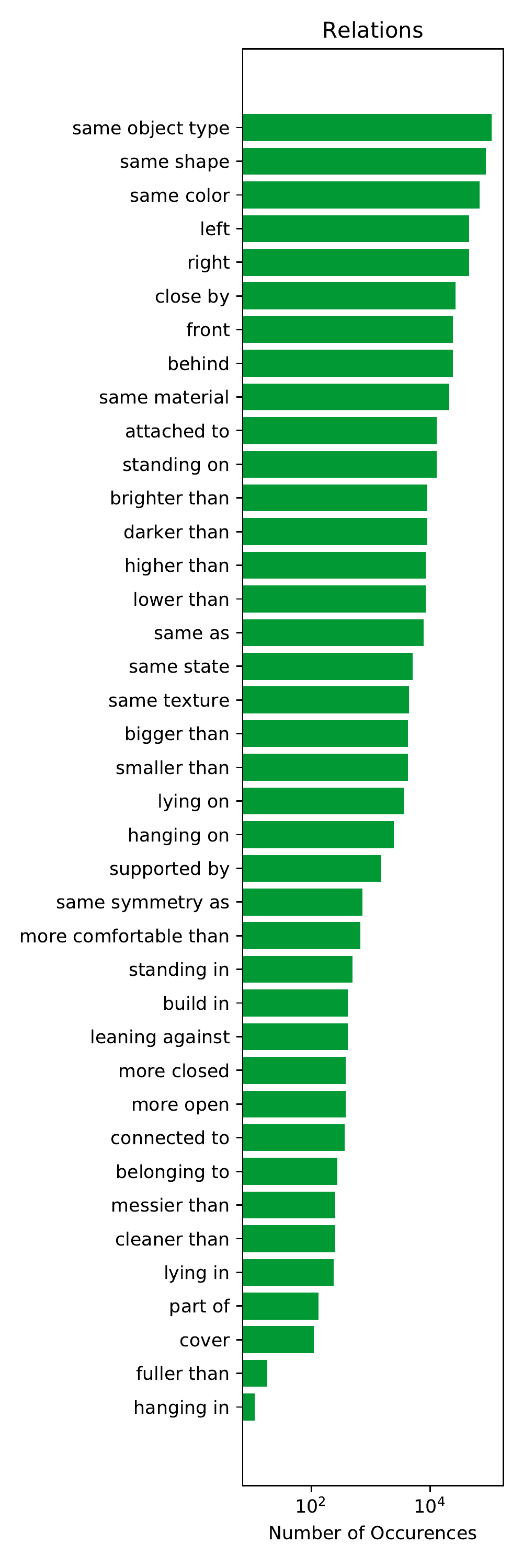}
\end{center}
   \caption{Predicate classes, sorted by occurrence, presented in logarithmic scale}
\label{fig:stats_relationships}
\end{figure}

\begin{figure}[t]
\begin{center}
    \includegraphics[height=0.9\textheight]{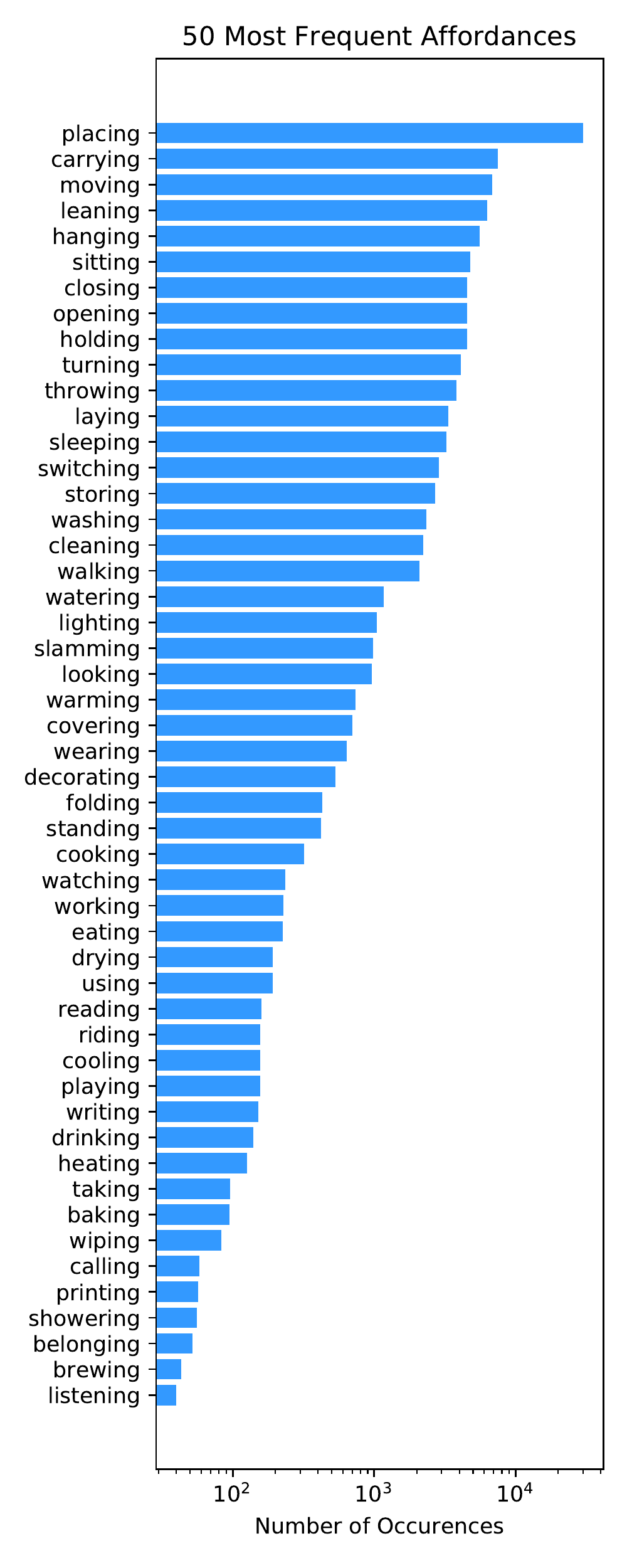}
\end{center}
   \caption{Simplified affordances in the \nameDataset{} dataset, sorted by occurrence, presented in logarithmic scale. Please note that for visualization purposes nouns and prepositions are removed such that \eg \rel{hanging in} or \rel{hanging on} are combined into \rel{hanging}.}
\label{fig:stats_affordances}
\end{figure}

\begin{figure*}[t]
\vspace*{-1cm}
\begin{center}
    \includegraphics[height=\textheight]{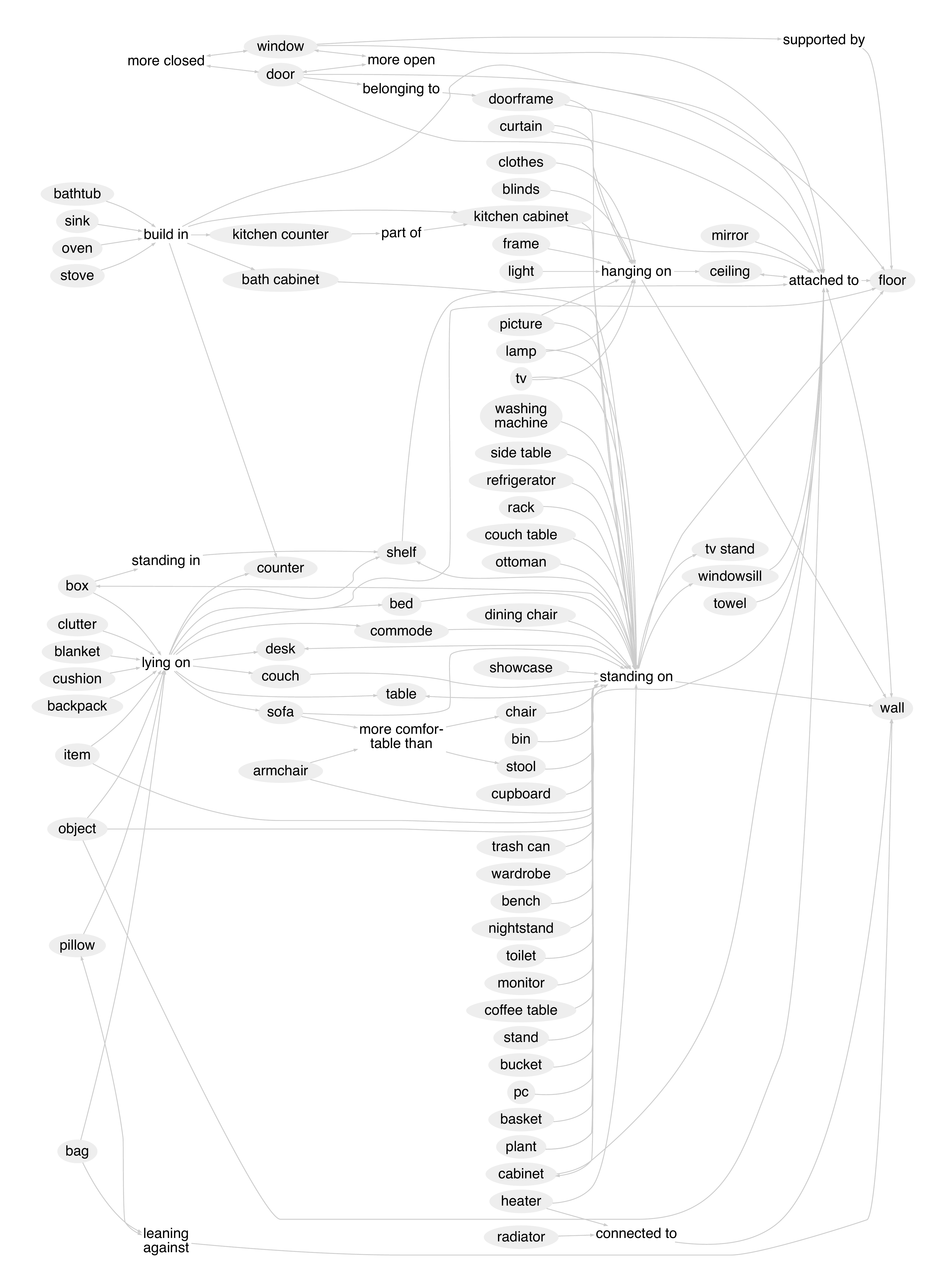}
\end{center}
   \caption{Simplified most frequent semantic \rel{(subject, predicate, object)} tuples with more than 50 occurrences in the \nameDataset{} dataset. Please note that for simplification purposes proximity relationships and most comparative relationships are filtered in this graph.}
    \label{fig:semantic_graph_rio}
\end{figure*}

\subsection{Graph Prediction}
\label{sec:graph_pred_sup}

\definecolor{mygray}{gray}{0.6}

\paragraph{Implementation details}
For the feature extraction of nodes and edges, we adopt two standard PointNet architectures. The input points to ObjPointNet have three channels $(x,y,z)$, while the RelPointNet inputs have four $(x,y,z,\mathcal{M}_{i,j})$. The size of the final features (per node and per edge) is $256$. For the baseline model, the object and relationship predictors have namely three fully connected layers followed by batch norm and relu. The GCN encompasses $l=5$ layers, where $g_1(\cdot)$ and $g_2(\cdot)$ are composed of a linear layer followed by a relu activation. The class prediction MLPs consists of 3 linear layers with batch normalization and relu activation. We set $\lambda_{obj}=0.1$. For the per-class binary classification loss, $\alpha_t$ is set to $0.25$.  We use an Adam optimizer with a learning rate of $10^{-4}$. 

\paragraph{Data Processing} Since our provided scene graphs are extremely dense (see statistics) a pre-processing and filtering of the ground truth graph data was required. We split the original graphs into smaller subgraphs of 4--9 nodes. We further consider only a subset of the relationships. Similarly, object instances with uncommon classes are filtered. In summary, in our experiments we use 160 different object classes and 26 relationships. For reproducibility our splits are made publicly available.

\begin{figure*}[h!]
    \centering
    \includegraphics[width=\linewidth]{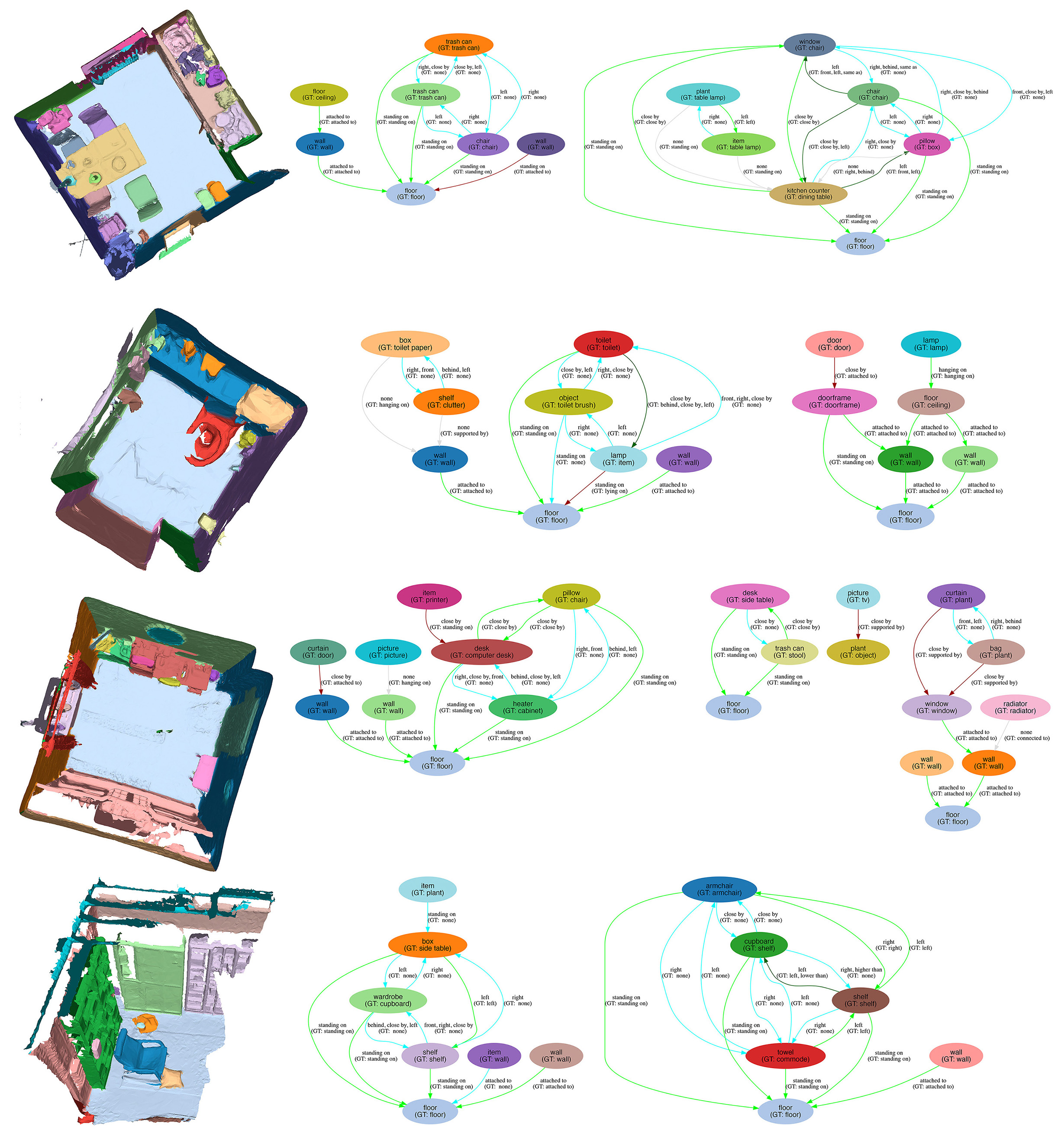}
    \caption{Qualitative results of our scene graph prediction model (best viewed in the digital file).}
    \label{fig:graph_pred_3D}
\end{figure*}

\begin{figure*}[h]
\begin{center}
   \includegraphics[width=\linewidth]{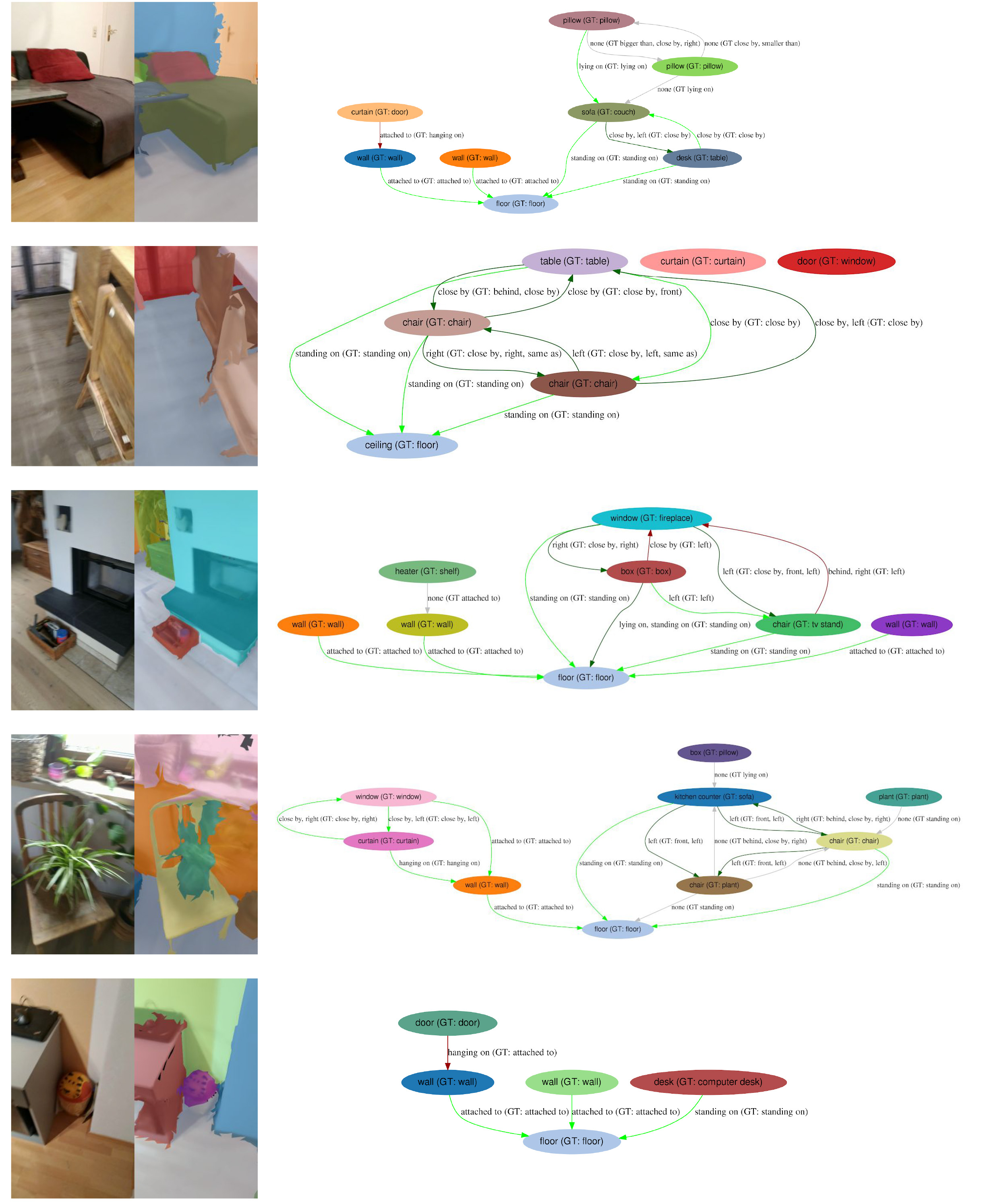}
\end{center}
   \caption{Qualitative results of our scene graph prediction model rendered to 2D (best viewed in the digital file).}
\label{fig:graph_pred_suppl_v1}
\end{figure*}

\begin{figure*}[h]
\begin{center}
   \includegraphics[width=1.05\linewidth]{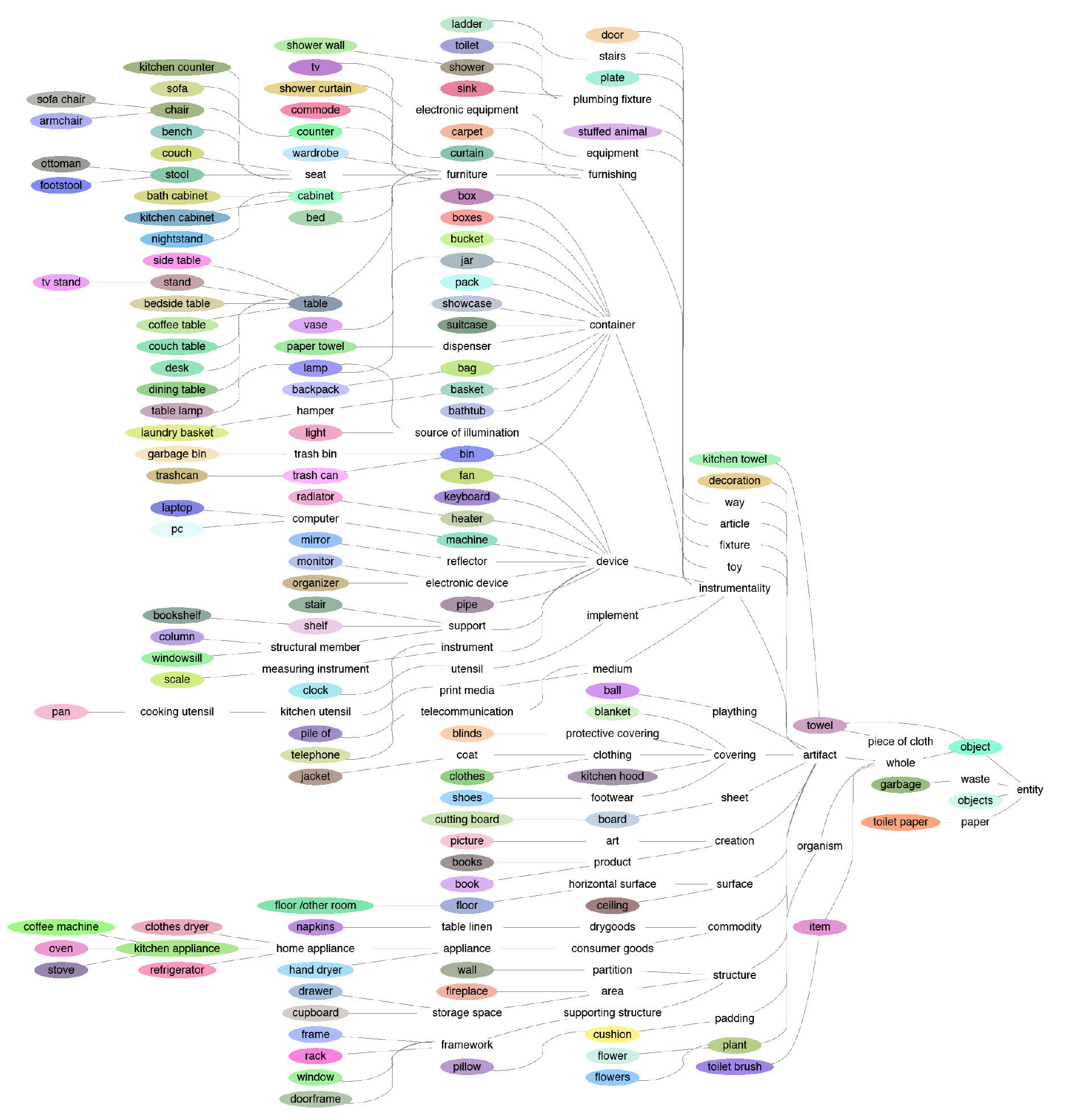}
\end{center}
   \caption{Graphical visualization of the hypernym / lexical relationships on a (bigger) subset of classes from 3RScan (see Fig. 3 in main paper).}
\label{fig:word_rel_graph}
\end{figure*}

\begin{figure*}[h]
\begin{center}
   \includegraphics[width=\linewidth]{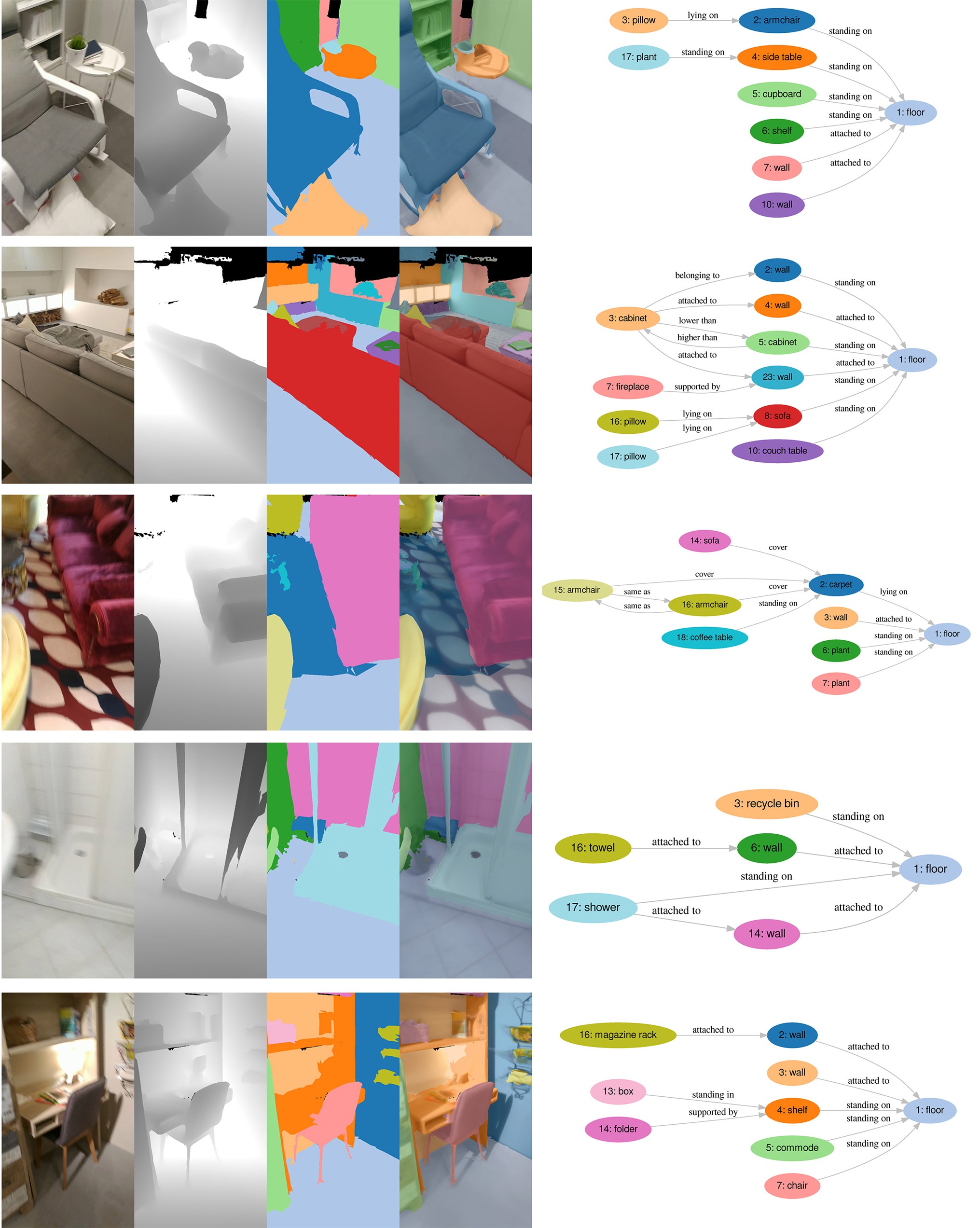}
\end{center}
   \vspace{-0.25cm}
   \caption{Rendered 2D graphs with a small subset of the relationships from our newly created 3D semantic scene graph dataset \nameDataset{ }. From left to right: RGB image, rendered depth, rendered dense semantic instance segmentation, dense semantic instance segmentation on textured model, 2D semantic scene graph.}
\label{fig:graph2D}
\end{figure*}

\paragraph{Qualitative Results} Fig. \ref{fig:graph_pred_3D} and \ref{fig:graph_pred_suppl_v1} show more qualitative semantic scene graph results using our proposed network architecture. Please note that Fig. \ref{fig:graph_pred_suppl_v1} shows rendered 2D subgraphs. Miss-classifications are reasonable (\rel{desk} vs. \rel{computer desk}, \rel{object} vs. \rel{toilet brush} or \rel{picture} vs. \rel{tv}). \textcolor{green}{\rel{Bright green}} edges are correctly predicted relationships between two notes; \textcolor{ForestGreen}{\rel{dark green}} edges are partially correct (a subset of the edges is correctly predicted and the rest is either missing or miss-classified), \textcolor{ProcessBlue}{\rel{bright blue}} edges are false positives (and often semantically correct), \textcolor{BrickRed}{\rel{red}} edges are completely miss-classified, while \textcolor{mygray}{\rel{gray}} edges are missing in the prediction (predicted as \rel{none} while there exists an edge in the ground truth). In all edge and node predictions we show the prediction together with the ground truth data in brackets \eg \rel{shower curtain (GT: curtain)} in Fig. 6 (main paper) and Fig. \ref{fig:graph_pred_3D} and \ref{fig:graph_pred_suppl_v1}.

\paragraph{Class-Agnostic Instance Segmentation} -- the input of our graph prediction network -- is taken directly from the dense 3D ground truth instance segmentation from 3RScan. Since we only use the segmentation information (and not the class labels itself) we decided to name it \textit{class-agnostic} instance segmentation. In theory, every 3D geometric segmentation method that is able to segment separate instances could be used as the input of our method.

\subsection{Scene Retrieval}
\label{sec:retrieval}

\begin{figure*}[ht]
\begin{center}
   \includegraphics[width=0.7\linewidth]{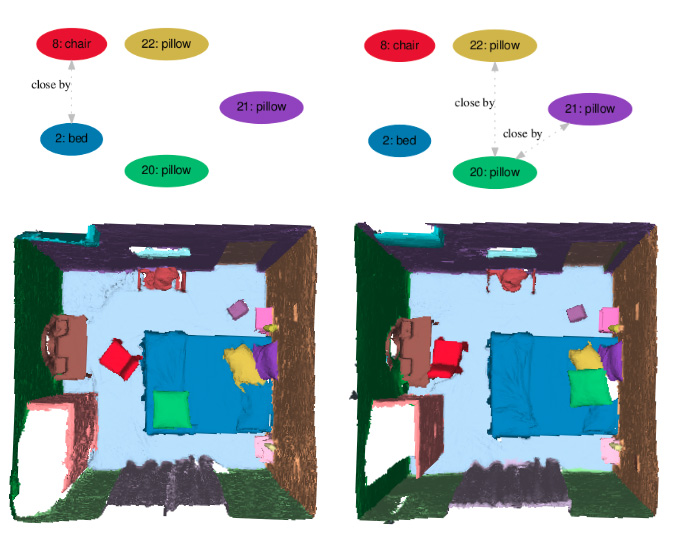}
   \includegraphics[width=0.7\linewidth]{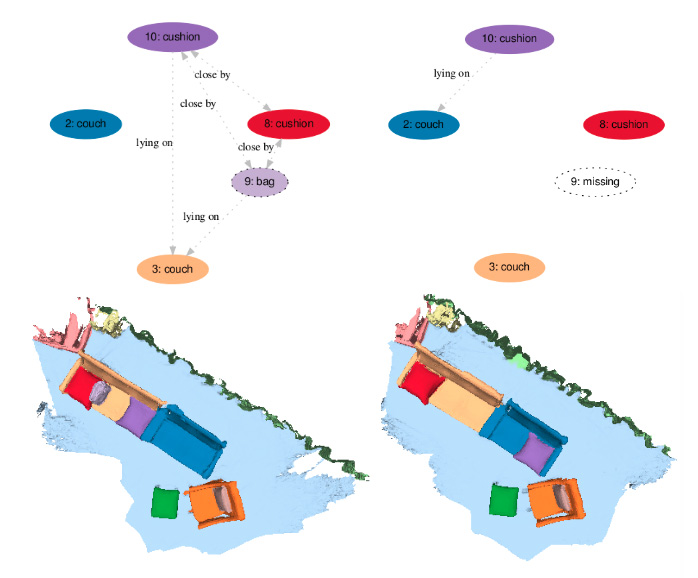}
\end{center}
   \caption{A byproduct of our scene retrieval is semantic change detection: Changed (added or removed) relationships and involved objects on two example scenes. Since we only show changes, all relationships \eg between pillows and the bed are not visualized.}
\label{fig:changes_1}
\end{figure*}

In Tbl. 3 of the main paper 3D-3D scene retrieval results are reported using ground truth graphs. While this gives us an upper bound for the scene retrieval task using 3D semantic scene graphs it also detects semantic changes. This interesting side effect is visualized in Fig. \ref{fig:changes_1}. Since only the parts of the graphs (nodes and edges) that could not be matched in the retrieval are visualized, it is easy to identify changes. In the upper example: 1.) the chair (8) pushed in a direction away from the bed (not \rel{close by} anymore) and 2.) a pillow (20) was moved closer to pillow (22) and pillow (21). In the lower example 1.) a bag was added (\rel{close by} cushion (8) and cushion (10)) and 2.) the purple cushion (10) was moved from couch (3) to couch (2). In these terms, the amount of change in a scene is the reverse of its ground truth similarity. 

\end{document}